\documentclass[preprint,12pt,
]{elsarticle}
\usepackage[a4paper, total={6.5in, 10in}]{geometry}
\usepackage{amssymb}
\usepackage{amsmath}
\usepackage{subfig}
\usepackage{multirow}
\usepackage{lineno}
\usepackage{hyperref}
\usepackage{tabularx}
\journal{Knowledge-Based Systems}
\begin{document}
	
\begin{frontmatter}
\title{Evolution of Image Segmentation using Deep Convolutional Neural Network: A Survey}
\author[add1]{Farhana Sultana}
\author[add1]{Abu Sufian \corref{cor1}}
\ead{sufian.csa@gmail.com}
\author[add2]{Paramartha Dutta}
\cortext[cor1]{Corresponding author}
\address[add1]{Department of Computer Science, University of Gour Banga, India}
\address[add2]{Department of Computer and System Sciences, Visva-Bharati University,  India }

\begin{abstract}
From the autonomous car driving to medical diagnosis, the requirement of the task of image segmentation is everywhere. Segmentation of an image is one of the indispensable tasks in computer vision. This task is comparatively complicated than other vision tasks as it needs low-level spatial information. Basically, image segmentation can be of two types: semantic segmentation and instance segmentation. The combined version of these two basic tasks is known as panoptic segmentation. In the recent era, the success of deep convolutional neural networks (CNN) has influenced the field of segmentation greatly and gave us various successful models to date. In this survey, we are going to take a glance at the evolution of both semantic and instance segmentation work based on CNN. We have also specified comparative architectural details of some state-of-the-art models and discuss their training details to present a lucid understanding of hyper-parameter tuning of those models. We have also drawn a comparison among the performance of those models on different datasets. Lastly, we have given a glimpse of some state-of-the-art panoptic segmentation models.
\end{abstract}


\end{frontmatter}

\section{Introduction}
\label{Sec:1}
We are living in the era of artificial intelligence (AI) and the advancement of deep learning is fueling AI to spread over rapidly \cite{deepLearning_nature15}, \cite{deeplearningBook16}. Among different deep learning models, convolutional neural network(CNN) \cite{wu2017introduction, o2015introduction, Ghosh2020} has shown outstanding performance in different high level computer vision task such as image classification  \cite{Krizhevsky2012, zeiler14, simonyan14, szegedy15, he16, lecun98, lecun90, Sabour2017, Hu17, Sultana2018},  object detection \cite{Girshick2016, He2014, Girshick2015, Ren2015, He2017, Dai2016, Redmon2015, Liu2016, Redmon2017, Zhang2017, Lin2017, Peng2017, WANG201962, Sultana2019} etc. Though the advent and success of AlexNet \cite{Krizhevsky2012} turned the field of computer vision towards CNN from traditional machine learning algorithms. But the concept of CNN was not a new one. It started from the discovery of Hubel and Wiesel \cite{hubel68} which explained that there are simple and complex neurons in the primary visual cortex and the visual processing always starts with simple structures such as oriented edges. Inspired by this idea, David Marr gave us the next insight that vision is hierarchical \cite{Marr1982}. Kunihiko Fukushima was deeply inspired by the work of Hubel and Wiesel and  built a multi-layered neural network called  Neocognitron \cite{Fukushima1980} using simple and complex neurons. It was able to recognize patterns in images and was spatial invariant. In 1989, Yann LeCun turned the theoretical idea of Neocognitron into a practical one called  LeNet-5 \cite{LeCun1989}. LeNet-5 was the first CNN developed for recognizing handwritten digits. LeCun et al. used back propagation \cite{LeCun1988}\cite{lecun98} algorithm to train his CNN. The invention of LeNet-5 paved the way for the continuous success of CNN in various high-level computer vision tasks as well as motivated researchers to explore the capabilities of such networks for pixel-level classification problems like image segmentation. The key advantage of CNN over traditional machine learning methods is the ability to learn appropriate feature representations for the problem at hand in an end-to-end training fashion instead of using hand-crafted features that require domain expertise \cite{GU2018354}.

  Applications of image segmentation are very vast. From the autonomous car driving \cite{Brabandere_2017_CVPR_Workshops} to medical diagnosis \cite{RIZWANIHAQUE2020100297, SONG201940}, the requirement of the task of image segmentation is everywhere. Therefore, in this article, we have tried to give a survey of different image segmentation models. This survey study has covered recent CNN-based state-of-the-art. Mainly semantic segmentation and instance segmentation of an image are discussed.  Herein, we have described comparative architectural details of notable different state-of-the-art image segmentation models. Also, different aspects of those models are presented in tabular form  for clear understanding. In addition, we have given a glimpse of recent state-of-the-art panoptic segmentation models.
 
 \subsection{Contributions of this paper}
    \begin{itemize}
    \item Gives taxonomy and survey of the evolution of CNN based image segmentation.
	\item Explores elaborately some CNN based popular state-of-the-art segmentation models. 
	\item Compares training details of those models to have a clear view of hyper-parameter tuning. 
	\item Compares the performance metrics of those state-of-the-art models on different datasets.
	\end{itemize}
\subsection{Organization of the Article} 
Starting from the introduction in section 1, the paper is organized as follows: 
In section 2, we have given background details of our work. In sections 3 and 4, semantic segmentation and instance segmentation works are discussed respectively with some subsections. In section 5, Panoptic segmentation is presented in brief. The paper is concluded in section 6. 
\section{Background Details}\label{seg}
\subsection{Why Convolutional Neural Networks?}
The computer vision has various tasks among them image segmentation as mentioned in section \ref{seg} is the focus of this article. Various researchers are addressing this task in different way using traditional machine learning algorithms like in \cite{Viola2001, Vaillant1994, Dollar2009} with the help various technique such as thresholding \cite{Sivakumar2014}, region growing \cite{Omez2007,Mary2012}, edge detection \cite{Huang2010, al2010,ma1997}, clustering \cite{Zheng2018, Mehmet1990, Kavitha2010, Ali2006, Galbiati2009, Franek2011, Alush2013, Yarkony2012}, super-pixel  \cite{Chang2012, XIe2019}, etc for years. Most of the successful works are based on handcrafted machine learning features such as HOG \cite{Dalal2005,chang2011, tuermer2013, gupta2014}, SIFT \cite{Burger2016, Akira2008}, etc. First of all, feature engineering needs domain expertise and the success of those machine learning-based models was slowed down around the era when deep learning was started to take over the world of computer vision. To give a outstanding performance, deep learning only needs data and it does not need any traditional handcrafted feature engineering techniques. Also, traditional machine learning algorithm can not adjust itself for a wrong prediction. On the other hand, deep learning has that capability to adapt itself according to the predicted result.  Among different deep learning algorithms, CNN got tremendous success in different fields of computer vision as well as grab the area of image segmentation \cite{srinivas2016taxonomy, chen2014semantic}.  
\subsection{Image Segmentation}
\label{seg}
\begin{figure}[htb]
	\centering
	\includegraphics[scale=0.55]{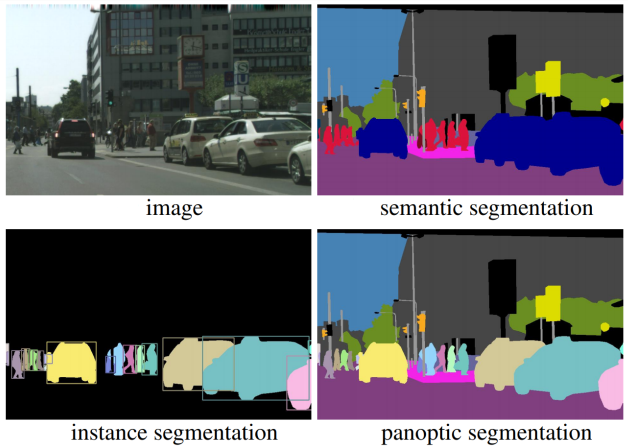}
	\caption{An example of different types of image segmentation. From \cite{kirillov2019panoptic}}.
	\label{all_seg}
\end{figure}

In computer vision, image segmentation is a way of segregating a digital image into multiple regions according to the different properties of pixels. Unlike classification and object detection, it is typically a low-level or pixel-level vision task as the spatial information of an image is very important for segmenting different regions semantically. Segmentation aims to extract meaningful information for easier analysis. In this case, the image pixels are labeled in such a way that every pixel in an image shares certain characteristics such as color, intensity, texture, etc. \cite{PAL19931277, zaitoun2015survey}. 
\begin{figure}[htb]
	\centering
	\includegraphics[scale=0.37]{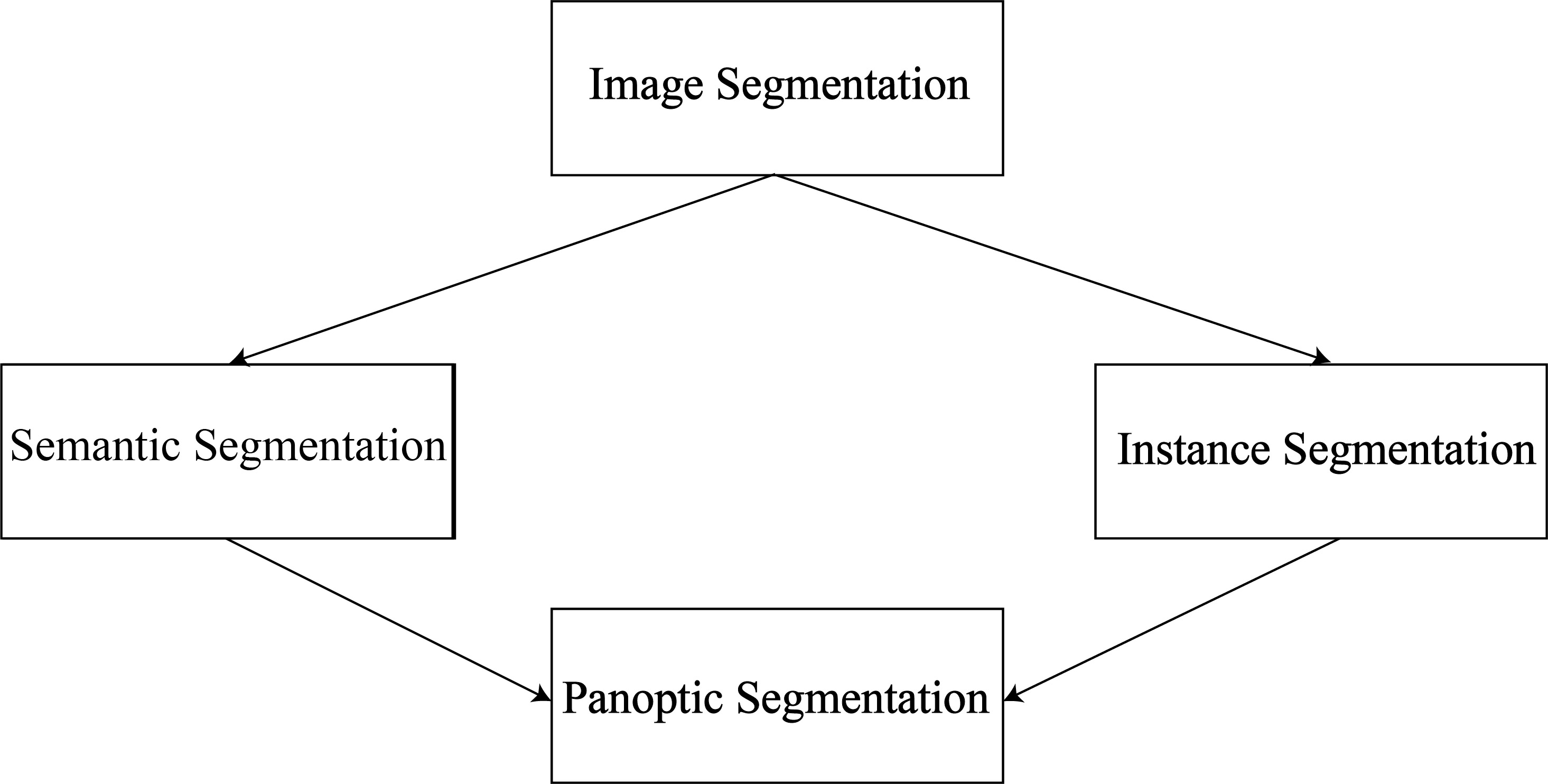}
	\caption{Different types of image segmentation}
	\label{types_seg}
\end{figure}  
Mainly, image segmentation is of two types: semantic segmentation and instance segmentation. Also, there is another type called panoptic segmentation\cite{Kirillov2018} which is the unified version of two basic segmentation processes. Figure \ref{types_seg} shows different types of segmentation and figure \ref{all_seg} shows the same with examples. In subsequent sections, we have elaborately discussed state-of-the-art of different CNN-based image segmentations.   

In addition, CNN is also used successfully for video object segmentation. In a study \cite{Caelles2017}, Caelles et al. have first used a Fully convolutional network for one-shot video object segmentation. In another study \cite{Oh2018}, the authors have used ResNet \cite{he16} based Siamese Encoder with Global Convolutional Block for video object segmentation. On the other hand Miao et al. have used a CNN based semantic segmentation network and proposed Memory Aggregation Network (MA-Net) \cite{miao2020memory} to handle interactive video object segmentation(iVOS). The authors of \cite{yang2020collaborative} has used a CNN based semantic segmentation network as a base network for Collaborative Video Object Segmentation by
Foreground-Background Integration(CFBI). These are some of CNN- based video segmentation models that got state-of-the-art results on various video segmentation datasets. Due to the scope and size of the article, we have not covered this topic in detail in the present article.
\section{Semantic Segmentation} \label{sem_seg}
\begin{figure}[htb]
	\centering
	\includegraphics[width=.9\linewidth]{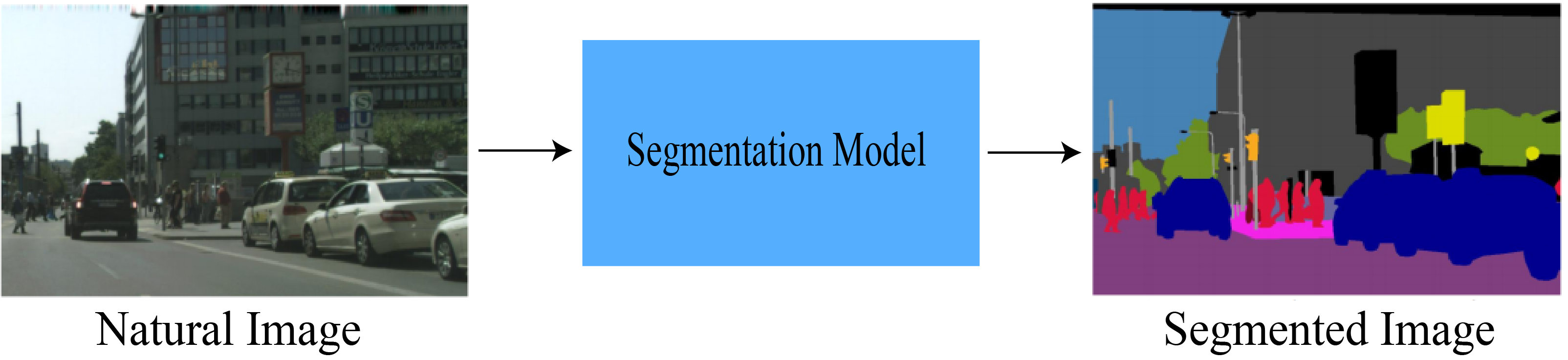}
	\caption{The process of semantic segmentation.}
	\label{fsem_seg}
\end{figure}

Semantic segmentation describes the process of associating each pixel of an image with a class label \cite{WEI2016234}. Figure \ref{fsem_seg} shows the black-box view of semantic segmentation. After the success of AlexNet in 2012,  we have got different successful semantic segmentation models based on CNN. In this section, we have present a survey of the evolution of CNN based semantic segmentation models. In addition, we have brought up here an elaborate exploration of some state-of-the-art models.
  
\subsection{Evolution of CNN based Semantic Segmentation Models}

 The application of CNN in semantic segmentation models has started with a huge diversity. In study \cite{Farabet2013}, the authors have used multi-scale CNN for scene labeling and achieved state-of-the-art results in the Sift flow \cite{Liu2011}, the Bercelona dataset \cite{Tighe2010} and the Standford background dataset \cite{Gould2009}. R-CNN \cite{Girshick2014} used selective search \cite{Uijlings2013} algorithm to extract region proposals first and then applied CNN upon each proposal for PASCAL VOC semantic segmentation challenge \cite{Everingham2007}. R-CNN achieved record result over second order pooling ($O_2 P$) \cite{Carreira2012} which was a leading hand-engineered semantic segmentation system at that time. At the same time, Gupta et al. \cite{gupta2014} used CNN along with geocentric embedding on RGB-D images for semantic segmentation. 
 
  Among different CNN based semantic segmentation models, Fully Convolutional Network(FCN) \cite{Long2017}, as discussed in subsection \ref{ss_fcn}, gained the maximum attention and an FCN based semantic segmentation model trend has emerged. To retain the spatial information of an image, FCN based models removed fully connected layers of traditional CNN.  In studies \cite{Mostajabi2014} and \cite{Szegedy2014}, the authors have used contextual features and achieved state of the art performance. Recently, in \cite{YangHu2019}, the authors have used fully convolutional two stream fusion network for interactive image segmentation.
   
  Chen et al aggregate `atrous' algorithm and conditional random (CRF) field in semantic segmentation and proposed DeepLab \cite{chen2014} as discussed in subsection \ref{ss_dialation}. Later the authors have incorporated `Atrous Special Pyramid Pooling (ASPP)'  in DeepLabv2 \cite{Chen2016}. DeepLabv3 \cite{Chen2017} has gone further and used a cascaded deep ASPP module to incorporate multiple contexts. All three versions of DeepLab have achieved good results.
  
   Deconvnet \cite{Noh2015} used convolutional network followed by hierarchically opposite de-convolutional network for semantic segmentation as discussed in section \ref{ss_tdbu}. Ronneberger et al used a U-shaped network called U-Net \cite{Ronneberger2015} which has a contracting and an expansive pathway to approach semantic segmentation. Contracting path extracts feature maps and reduces spatial information as a traditional convolution network. Expansive pathway takes the contracted feature map as input and apply an up-convolution. 
   Section \ref{ss_tdbu} discussed the model in more detail. Recently, in \cite{Nabil2020}, the authors have used U-Net with $multiRes$ block for multimodal biomedical image segmentation and got better result than using classical U-Net. SegNet \cite{Badrinarayanan2015} is a encoder-decoder network for semantic segmentation. The encoder is a basic VGG16 network excluding FC layers. The decoder is identical to encoder but the layers are hierarchically opposite.   
   SegNet is discussed in section \ref{ss_tdbu}. The basic architectural intuition of U-Net, Deconvnet, and SegNet are similar except some individual modifications. The second half of those architectures is the mirror image of the first half.
 
  Liu et al. mixed the essence of global average pooling and L2 normalization layer in FCN \cite{Long2017} architecture, and proposed ParseNet \cite{Liu2015} to achieve state of the art result in various datasets.  Zhao et al. proposed Pyramid Scene Parsing Network(PSPNet) \cite{Zhao2016}. They have used Pyramid Pooling Module on top of the last extracted feature map to incorporate global contextual information for better segmentation. Peng et al. used the idea of global convolution using a large kernel to apply the advantage of both local and global features. Pyramid Attention Network (PAN) \cite{LiH2018}, ParseNet \cite{Liu2015}, PSPNet\cite{Zhao2016} and GCN\cite{PengC2017} have used global context information with local feature to have better segmentation. Sections \ref{ss_gc} and \ref{ss_rfmc} will discuss those models in detail.
  
   Fully convolutional DenseNet \cite{he16} is used to address semantic segmentation in \cite{Brahimi2019, Towaki2019}. DeepUNet \cite{ LiR2017}, a ResNet based FCN, used to segment sea land.  At the same time, ENet\cite{Paszke16}, ICNet\cite{zhao2018} are used as real-time semantic segmentation models for the autonomous vehicles.   
   Some recent works \cite{Chieh2018, liu2019auto, Sun2019} have used combination of encoder-decoder architecture and dilated convolution for better segmentation. Transfer learning or domain adaption also uses for semantic segmentation \cite{ZHANG2019105444}. Kirillov et al.\cite{he27pointrend} used point-based rendering in DeepLabV3\cite{Chen2017} and in semanticFPN to produce state-of-the-art semantic segmentation models. 
  
  Researcher from different field of deep learning has also infused CNN to address semantic segmentation. In study\cite{luc2016semantic}, the authors have trained CNN along with adversial network. Luo et al. have also used CNN as generator and discriminator in a adversial network and proposed Category level Advisory Network(CLAN)\cite{Luo2018}. In \cite{Luo2019S}, The authors have used same configuration as CLAN with information bottleneck for domain adaptive semantic segmentation and proposed Information Bottlenecked Adversarial Network(IBAN) and  Significance-aware Information Bottlenecked Adversarial Network (SIBAN). In another study \cite{LuoY2018}, the authors have used CNN based adversial network named Macro-Micro Adversila Network (MMAN) for human parsing.  
  
  Some researcher have used CNN models for attention based image segmentation. Wang et al. used Non-local Neural Network in \cite{Wang2018}. Huang et al. used DeepLab for feature map extraction and then the feature maps are fed into recurrent criss cross attention module \cite{Huang2019} for semantic segmentation. In another study \cite{ZhaoH2018}, The authors have aggregated long range contextual information in convolutional feature map using global attention network to address scene parsing. In \cite{Fu2019}, the authors have used dual attention network in combination of CNN for scene segmentation.

\subsection{Some popular state-of-the-art semantic segmentation models}\label{ss_model} 
In this section, we are going to explore architectural details of some state of the art CNN based semantic segmentation models in detail. The models are categorized on the basis of the most impotant feature used. At the end of each categorical discussion, we have also briefly discussed the advantages and weaknesses of a particular model category.
\subsubsection{Based on Fully Convolutional Network:} \label{ss_fcn}
\textbf{FCN:}
Long et al. proposed the idea of Fully Convolutional Network(FCN) \cite{Long2017} to address the semantic segmentation task. They have used AlexNet\cite{Krizhevsky2012}, VGGNet\cite{simonyan14} and GoogleNet\cite{szegedy15}(all three pre-trained on ILSVRC \cite{ILSVRC15} data) as base models. They transferred these models from classifiers to dense FCN by substituting fully connected layers with $1\times1$ convolutional layers and append a  $1\times1$ convolution with channel dimension 21 to predict scores for each of the 20 PASCAL VOC \cite{Everingham2010} classes and background class. 
\begin{figure}[htb]
	\centering
	\includegraphics[width=.98\linewidth]{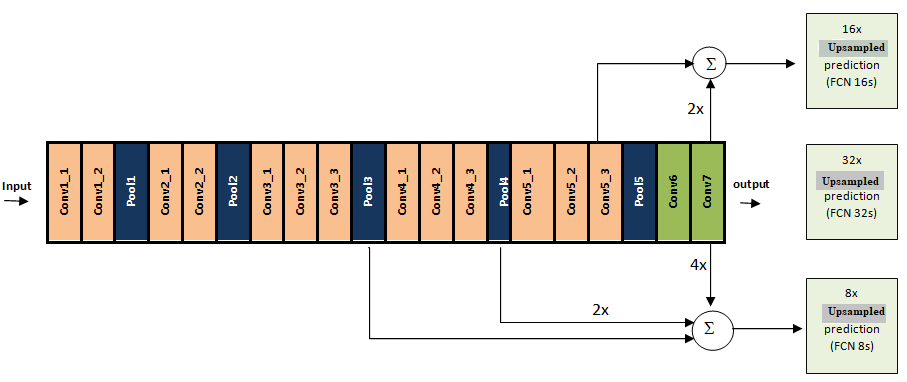}
	\caption{Architecture of FCN32s, FCN16s, FCN8s}
	\label{fcn}
\end{figure}  

This process produces a class presence heat map in low resolution. The authors have experienced that among FCN-AlexNet, FCN-VGG16 and FCN-GoogLeNet, FCN-VGG16 gave the highest accuracy on PASCAL VOC 2011 validation dataset. So, they choose the FCN-VGG16 network for further experiments. As the network produces coarse output locations, the authors used bilinear interpolation to upsample the coarse output  $32\times$ to make it pixel dense. But this upsampling was not enough for fine-grained segmentation. So they have used skip connection\cite{Bishop2006} to combine the final prediction layer and feature-rich lower layers of VGG16 and call this combination as $deep$ $jet$. Figure \ref{fcn} shows different $deep$ $ jet$s : FCN-16s and FCN-8s and FCN-32s. Among them, FCN-8s gave the best result in PASCAL VOC 2011 \& 2012 \cite{Everingham2010} test dataset and FCN-16s gave the best result on both NYUDv2 \cite{Silberman2012} \& SIFT Flow \cite{Liu2011} datasets.

   Major changes in FCN which helped the model to achieve state of the art result are the base model VGG16, bipolar interpolation technique for up-sampling the final feature map and skip connection for combining low layer and high layer features in the final layer for fine-grained semantic segmentation. 
   
   FCN has used only local information for semantic segmentation but only local information makes semantic segmentation quite ambiguous as it looses global semantic context of the image. To reduce ambiguity contextual information from the whole image is much helpful. 

\subsubsection{Based on Dialtation/Atrous convolution:} \label{ss_dialation}
\begin{figure}[htb]
	\centering
	\includegraphics[width=.71\linewidth]{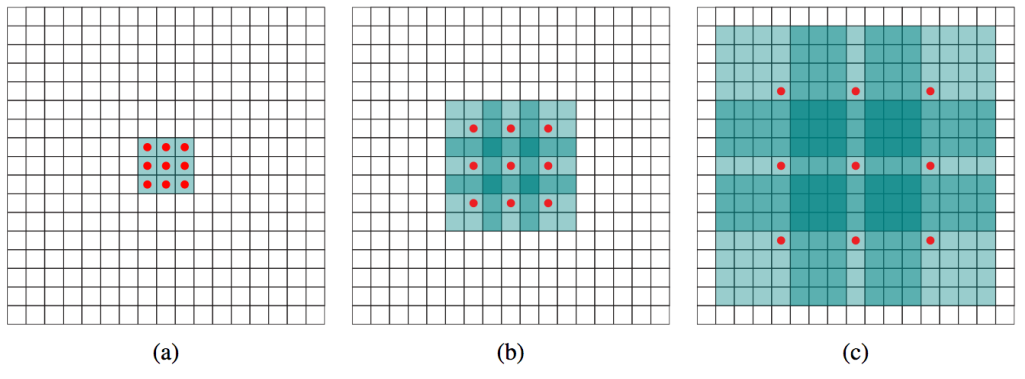}
	\caption{(a) 1-dialatednet with receptive field $3\times3$, (b) 2-dialatednet with receptive field $7\times7$ and (c)-dialatednet with receptive field $15\times15$. From \cite{YuK15}}.
	\label{Dialatednet}
\end{figure}

\textbf{Dialatednet:}
Traditional CNN, used for classification tasks, loses resolution in its way and it is not suitable for dense prediction. Yu and Koltun have introduced a modified version of traditional CNN, called dialated convolution or DialatedNet \cite{YuK15}, to accumulate multi-scale contextual information systematically for better segmentation without suffering the loss of resolution. DialatedNet is like a rectangular prism of convolutional layers, unlike conventional pyramidal CNN. Without losing any spatial information, it can support the exponential expansion of receptive fields as shown in figure \ref{Dialatednet}. 

\begin{figure*}[htb]
	\centering
	\subfloat[]{\includegraphics[width=6cm]{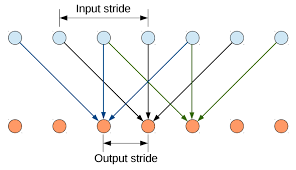}\label{hole}}
	\hspace{2mm}
	\subfloat[]{\includegraphics[width=6cm]{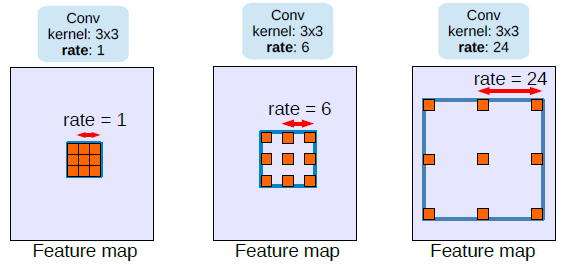}\label{hole2}}\\
	\caption{  Illustration of atrous algorithm (a) in 1-D, when kernel size=3, input-stride=2 and output-stride=1. From \cite{chen2014} and (b) in 2-D, when kernel size $3\times 3$, with rate 1, 6 and 24. From \cite{Chen2017}}
	\label{rcnn_sppnet}	
\end{figure*}

\textbf{DeepLab:}
Chen et al. has brought together methods from Deep Convolutional Neural Network(DCNN) and probabilistic graphical model. The authors have faced two technical difficulties in the application of DCNN to semantic segmentation: down sampling and spatial invariance. To handle the first problem, they have employed `atrous' \cite{mallat1999} algorithm for efficient dense computation of CNN. Figure \ref{hole} and \ref{hole2} shows atrous algorithm in 1-D and in 2-D.  To handle the second problem, they have applied a fully connected pairwise conditional random field (CRF) to capture fine details. In addition, the authors have reduced the size of the receptive field $6\times$ than the original VGG16 \cite{simonyan14} network to reduce the time consumption of the network and also used multi-scale prediction for better boundary localization.

Advantage of dilation based model is that it helps to retain spatial resolution of the image to produce dense prediction. But use of dilation convolution isolates image pixels from its global context which makes it prone to misclassification.

\subsubsection{Based on Top-down/Bottom-up approach:}\label{ss_tdbu} 
\begin{figure}[htb]
	\centering
	\includegraphics[width=.98\linewidth]{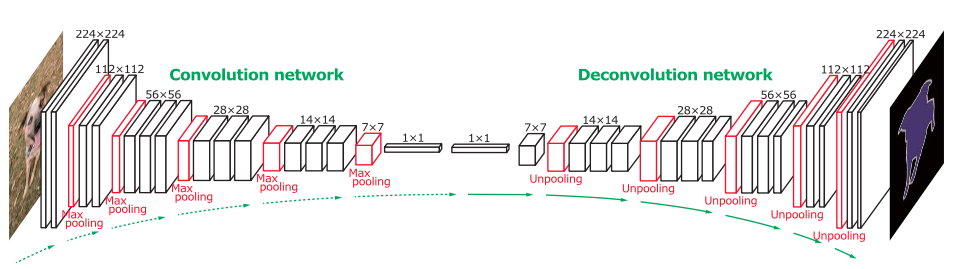}
	\caption{Convolution-Deconvolution  architecture of Decovnet. From \cite{Noh2015}}
	\label{Deconvnet}
\end{figure}

\textbf{Deconvnet:} Deconvnet \cite{Noh2015}, proposed by Noh et al., has a convolutional and de-convolutional network. The convolutional network is topologically identical with the first 13 convolution layers and 2 fully connected layers of VGG16\cite{simonyan14} except for the final classification layer.  As in VGG16, pooling and rectification layers are also added after some of the convolutional layers. The De-convolutional network is identical to the convolutional network but hierarchically opposite. It also has multiple series of deconvolution, un-pooling and rectification layers. All the layers of convolutional and de-convolutional network extract feature maps except the last layer of the de-convolutional network which generates pixel-wise class probability maps of the same size as the input image. In the deconvolutional network, the authors have applied unpooling which is the reverse operation of the pooling operation of the convolutional networks to reconstruct the original size of activation. Following \cite{zeiler14} and \cite{Zeiler2011}, unpooling is done using max-pooling indices which are stored at the time of convolution operation in the convolutional network. To densify enlarged but sparse un-pooled feature maps, convolution like operation is done using multiple learned filters by associating single input activation with multiple outputs. Unlike FCN, the authors applied their network on object proposals extracted from the input image and produced pixel-wise prediction. Then they have aggregated outputs of all proposals to the original image space for segmentation of the whole image. This instance wise segmentation approach handles multi-scale objects with fine detail and also reduces training complexity as well as memory consumption for training. To handle the internal covariate shift in the network, the authors have added batch normalization \cite{Bjorck2018} layer on top of convolutional and de-convolutional layers. The architecture of Deconvnet is shown in figure \ref{Deconvnet}.

\begin{figure}[htb]
	\centering
	\includegraphics[scale=0.23]{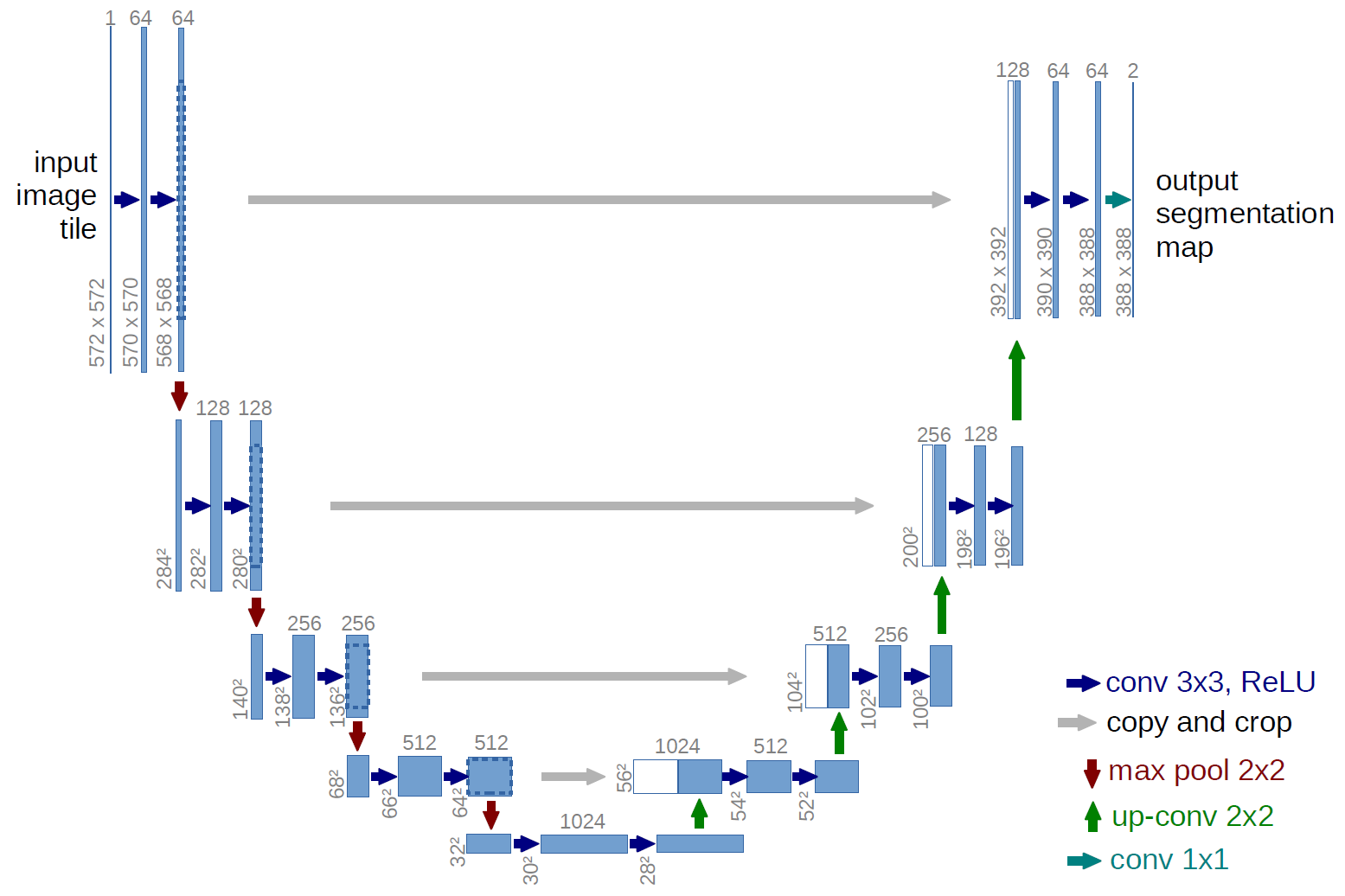}
	\caption{Contracting and expansive architecture of U-Net. From \cite{Ronneberger2015}}
	\label{u_net}
\end{figure}

\textbf{U-Net:} U-Net \cite{Ronneberger2015} is a U-shaped semantic segmentation network which has a contracting path and an expansive path. Every step of the contracting path consists of two consecutive $3\times3$ convolutions followed by ReLU nonlinearity and max-pooling using $2\times2$ window with stride 2. During the contraction, the feature information is increased while spatial information is decreased. On the other hand, every step of the expansive path consists of up-sampling of feature map followed by a $2\times2$ up-convolution. This reduces the feature map size by a factor of 2. Then the reduced feature map is concatenated with the corresponding cropped feature map from the contracting path. Then two consecutive $3\times3$ convolution operations are applied followed by ReLU nonlinearity. In this way, the expansive pathway combines the features and spatial information for precise segmentation. The architecture of U-Net is shown in figure \ref{u_net}.

\begin{figure}[htb]
	\centering
	\includegraphics[width=.92\linewidth]{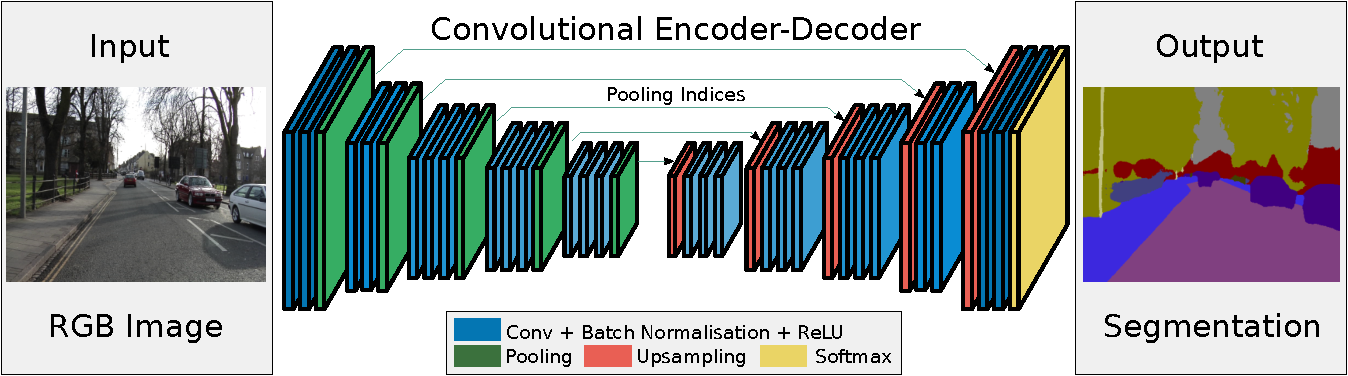}
	\caption{Encoder-decoder architecture of SegNet. From \cite{Badrinarayanan2015}}
	\label{SegNet}
\end{figure}

\textbf{SegNet:} SegNet  \cite{Badrinarayanan2015} has encoder-decoder architecture followed by a final pixel-wise classification layer. The encoder network has 13 convolutional layers as in VGG16 \cite{simonyan14} and the corresponding decoder part also has 13 de-convolutional layers. The authors did not use fully connected layers of VGG16 to retain the resolution in the deepest layer and it reduces the number of parameters from 134M to 14.7M. In each layer in the encoder network, a convolutional operation is performed using a filter bank to produce feature maps. Then, to reduce internal covariate shift the authors have used batch normalization \cite{Ioffe2015} \cite{Badrinarayanan2015(BN)} followed by ReLU \cite{agarap2018} nonlinearity operation. Resulting output feature maps are max-pooled using a $2\times2$ non-overlapping window with stride 2 followed by a sub-sampling operation by a factor of 2. A combination of max-pooling and sub-sampling operation achieves better classification accuracy but reduces the feature map size which leads to lossy image representation with blurred boundaries which is not ideal for segmentation purposes where boundary information is important. To retain boundary information in the encoder feature maps before sub-sampling, SegNet stores only the max-pooling indices for each encoder map. For semantic segmentation, the output image resolution should be the same as the input image. To achieve this, SegNet does up-sampling in its decoder using the stored max-pooling indices from the corresponding encoder feature map resulting high-resolution sparse feature map. To make the feature maps dense, the convolution operation is performed using a trainable decoder filter bank. Then the feature maps are batch normalized. The high-resolution output feature map produced form final decoder is fed into a trainable multi-class softmax classifier for pixel wise labeling. The architecture of SegNet is shown in figure \ref{SegNet}.

\begin{figure}[htbp]
	\centering
	\includegraphics[scale=0.35]{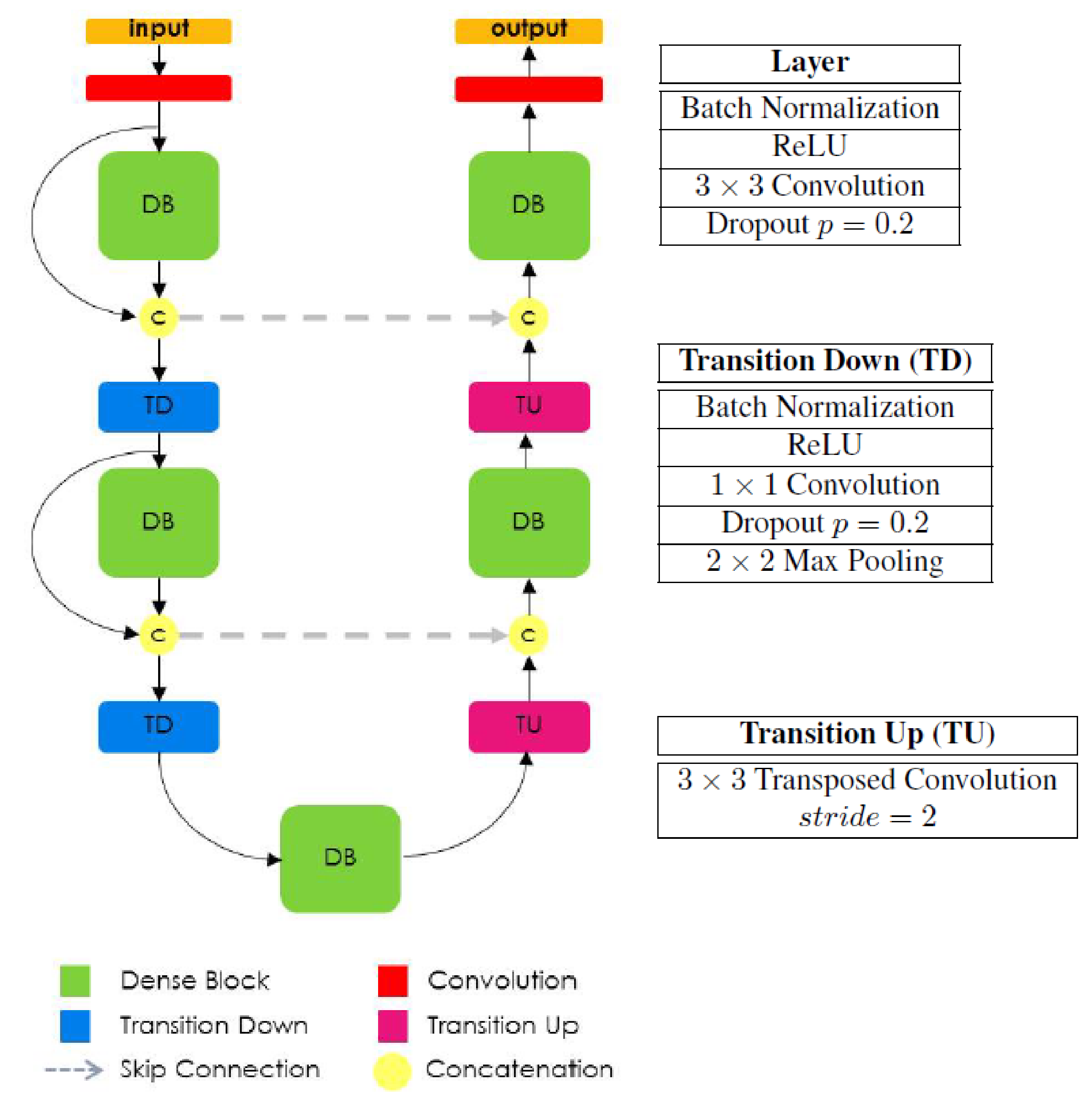}
	\caption{Architecture of Fully Convolutional DenseNet for semantic segmentation with some building blocks. From \cite{JegouDVRB16}}
	\label{FCDense}
\end{figure}

\textbf{FC-DenseNet:} DenseNet \cite{he16} is a CNN based classification network that contains only a down-sampling pathway for recognition.  J\'egou et al. \cite{JegouDVRB16} has extended DenseNet by adding an up-sampling pathway to regain the full resolution of the input image. To construct the up-sampling pathway, the authors followed the concept of FCN. They have referred the down-sampling operation of DenseNet as Transition Down (TD) and up-sampling operation in extended DenseNet as Transition UP (TU) as shown in figure \ref{FCDense}.  The rest of the convolutional layers follows the sequence of Batch Normalization, ReLU, $3\times3$ convolution and dropout of 0.2 as shown in the top right block in figure \ref{FCDense}. The up-sampling pathway used the sequence of dense block \cite{he16} instead of convolution operation of FCN and used transposed convolution as an up-sampling operation. The up-sampling feature maps are concatenated with the feature maps derived from corresponding layers of the down-sampling pathway. In figure \ref{FCDense}, these long skip connections are shown as yellow circle.

  As the upsampling rate of FCN based model in final layer is very high, it produces coarse output in final layer. So fine-grained semantic segmentation is not possible. On the other hand top-down/bottom-up approach based models used gradually increasing upsampling rate which leads to more accurate segmentation. But in this case the model also lacks incorporation of global contextual information.   

\subsubsection{Based on Global Context:} \label{ss_gc}

\begin{figure}[htb]
	\centering
	\includegraphics[width=.9\linewidth]{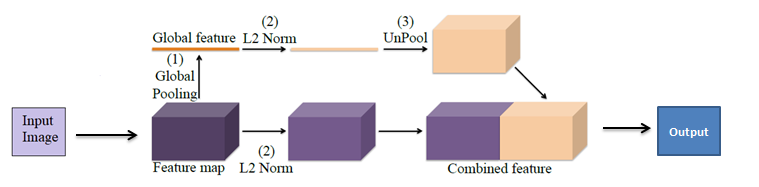}
	\caption{ParseNet Model Design \cite{Liu2015}}
	\label{ParseNet}
\end{figure}

\textbf{ParseNet:} \label{ss_parse}
Liu et al. proposed an end-to-end architecture called ParseNet \cite{Liu2015} which is an improvement of Fully Convolution Neural Network \cite{Long2017}. The authors have added global feature or global context information for better segmentation. In figure \ref{ParseNet}, the model description of ParseNet is shown. Till convolutional feature map extraction, the ParseNet is the same as FCN \cite{Long2017}. After that, the authors have used global average pooling to extract global contextual information. Then the pooled feature maps are un-pooled to get the same size as input feature maps. Now, the original feature maps and un-pooled feature maps are combined for predicting the final classification score. As the authors have combined two different feature maps from two different layers of the network, those feature maps would be different in scale and norm. To make the combination work, they have used two L2 normalization layers: one after global pooling and another after the original feature map extracted from FCN simultaneously. This network achieved state-of-the-art performance on ShiftFlow  \cite{Liu2011}, PASCAL-context \cite{mottaghi2014} and near the state of the art on PASCAL VOC 2012 dataset.

\textbf{GCN:}
Like  ParseNet, Global Convolution Network \cite{PengC2017} has also used global features along with local features to make the pixel-wise prediction more accurate. 
\begin{figure}[htbp]
	\centering
	\includegraphics[scale=0.20]{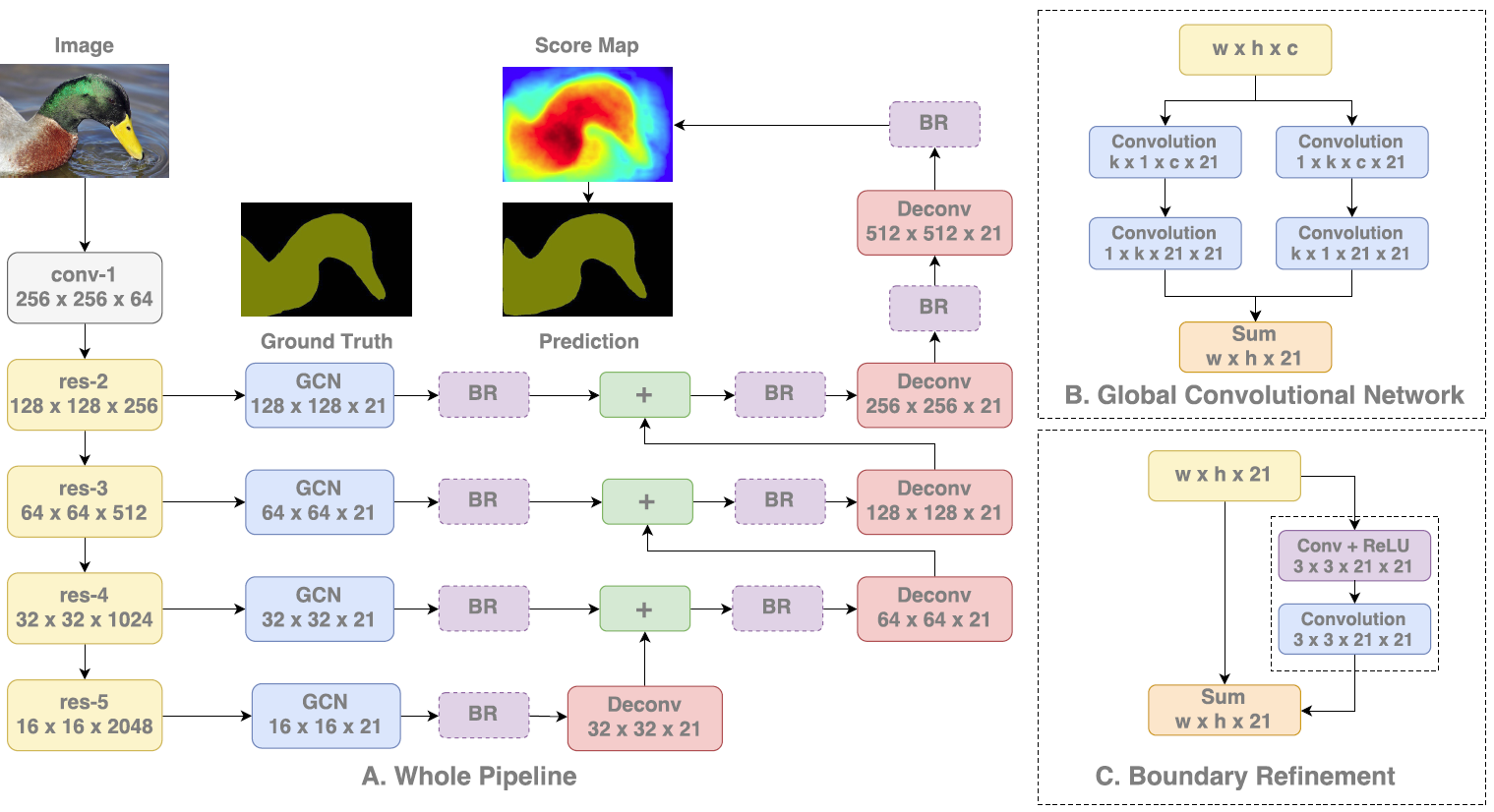}
	\caption{ Pipeline network of GCN. From \cite{PengC2017}}
	\label{GCN}
\end{figure}
The task of semantic segmentation is the combination of classification and localization tasks. These two tasks are contradictory in nature. The classification should be transformation invariant and localization should be transformation sensitive. Previous state-of-the-art models focused on localization more than classification. In GCN, the authors did not use any fully connected layers or global pooling layers to retain spatial information. On the other hand, they have used a large kernel size (global convolution) to make their network transformation invariant in the case of pixel-wise classification. To refine the boundary further the authors have used Boundary Refinement (BR) block.  As shown in figure \ref{GCN}, ResNet is used as a backbone. GCN module is inserted in the network followed by the BR module. Then score maps of lower resolution are up-sampled with a deconvolution layer, and then added up with higher ones to generate new score maps for final segmentation.

\begin{figure}[htbp]
	\centering
	\includegraphics[scale=0.45]{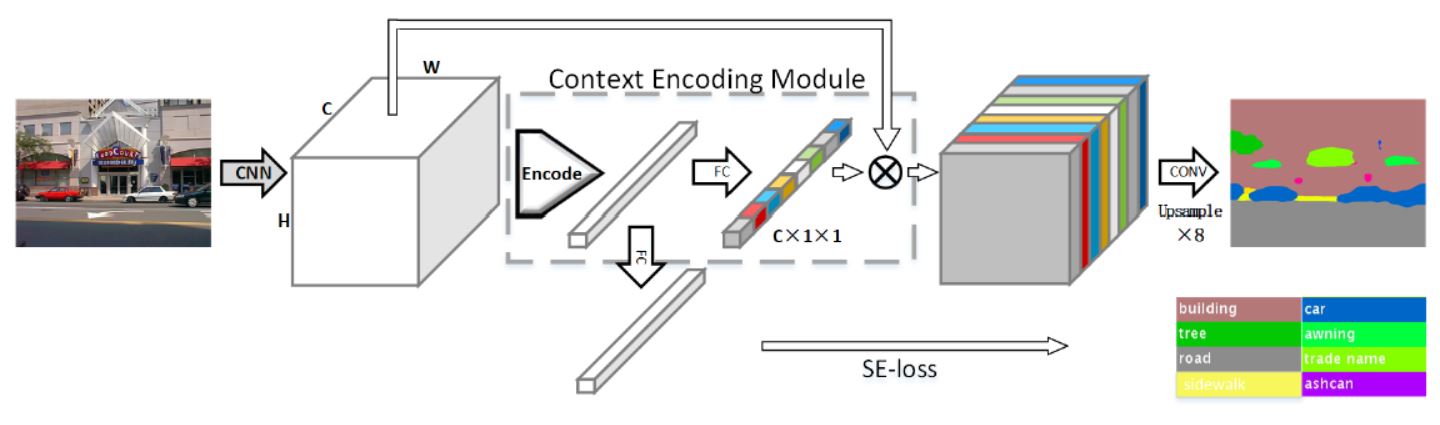}
	\caption{ EncNet from \cite{Zhang2018} (Notation: FC $\rightarrow$ Fully Convolutional Layer,Conv  $\rightarrow$ Convolutional Layer, Encode  $\rightarrow$ Encoding Layer and $\otimes$  $\rightarrow$ Channel wise multiplication )}
	\label{EncNet}
\end{figure}

\textbf{EncNet:} 
Zhang et al. have applied the idea of global context introducing novel context encoding module. The authors have used Semantic Encoding Loss(SE-loss) to help incorporation of global scene context information. This loss unit helps in regularizing the network training procedure in such a way that the network can predicts the presence of different category objects as well as learns the semantic context of an image. Using the above mentioned idea, the authors have proposed Context Encoding Network (EncNet)\cite{Zhang2018} as shown in figure \ref{EncNet}. The network contains a pre-trained Deep ResNet. On top of the ResNet, Context Encoding Model is used. The authors have used dialation strategy, multi GPU Batch Normalization and Memory efficient encoding layer to enhance the acuuracy of semantic segmetation.

 Though application of global convolution helps to improve accuracy but it lacks the scaling information of multi scale objects. 

\subsubsection{Based on receptive field enlargement and multi-scale context incorporation:}\label{ss_rfmc}
\begin{figure}[htb]
 	\centering
 	\includegraphics[width=.95\linewidth]{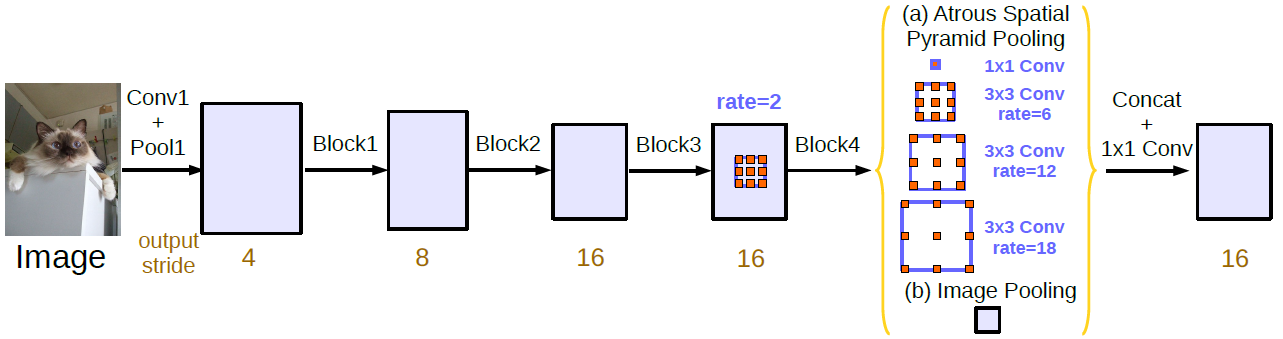}
 	\caption{Atrous Spatial Pooling Pyramid. From \cite{Chen2017}}
 	\label{aspp}
 \end{figure}

\textbf{DeepLabv2 and DeepLabV3:}
The authors of DeepLab modified their network using Atrous Special Pooling Pyramid (ASPP) to aggregate multi-scale features for better localization and proposed DeepLabv2 \cite{Chen2016}. Figure \ref{aspp} shows ASPP.  This architecture used both ResNet \cite{he16} and VGGNet\cite{simonyan14} as base network. 

In DeepLabv3\cite{Chen2017}, to incorporate multiple contexts in the network, the authors have used cascaded modules and have gone deeper especially with the ASPP module.

\textbf{PSPNet:} 
Pyramid Scene Parsing Network(PSPNet) \cite{Zhao2016}, proposed by Zhao et al.,  has used global contextual information for better segmentation. In this model, the authors have used Pyramid Pooling Module on top of the last feature map extracted using dilated FCN. In Pyramid Pooling Module, feature maps are pooled using 4 different scales corresponding to 4 different pyramid levels each with bin size $1\times1$, $2\times2$, $3\times3$ and $6\times6$. 
\begin{figure}[htbp]
\centering
\includegraphics[width=.96\linewidth]{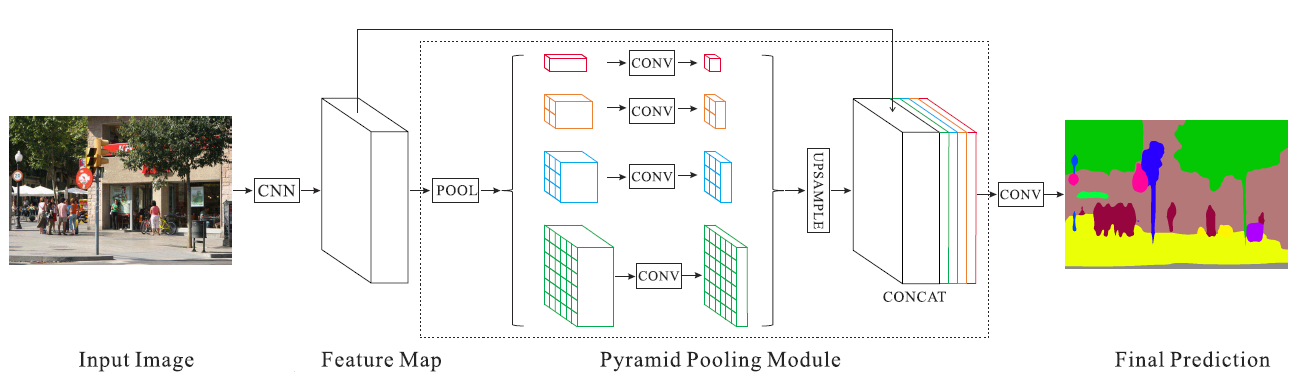}
\caption{PSPNet Model Design. From \cite{Zhao2016}}
\label{PSPNet}
\end{figure}
To reduce dimension, the pooled feature maps are convolved using $1\times1$ convolution layer. The output of the convolution layers are up-sampled and concatenated to the initial feature maps to finally combine the local and the global contextual information. Then, those output are again processed by a convolutional layer to generate the pixel-wise prediction. In this network, the pyramid pooling module observes the whole feature map in sub-regions with a different locations. In this way, the network understands a scene better which also leads to better semantic segmentation. In figure \ref{PSPNet}, the architecture of PSPNet is shown.

\begin{figure}[htbp]
	\centering
	\includegraphics[scale=0.40]{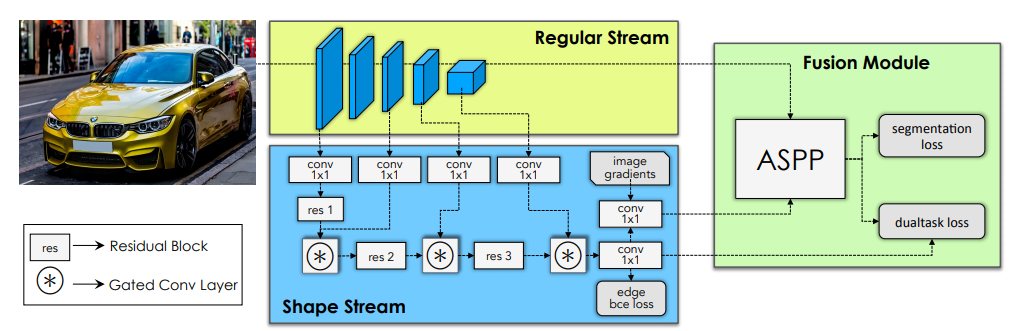}
	\caption{Architecture of Gated Shape CNN for semantic segmentation. From \cite{Towaki2019}}
	\label{GSCNN}
\end{figure}

\textbf{Gated-SCNN:} Takikawa et al. proposed Gated - Shape CNN(GSCNN) \cite{Towaki2019} for Semantic Segmentation. As shown in figure \ref{GSCNN}, GSCNN consists of two streams of networks: regular stream and shape stream. The regular stream is a classical CNN for processing semantic region information. Shape stream consists of multiple Gated Convolution Layer (GCL) which process boundary information of regions using low-level feature maps from the regular stream. Outputs of both streams are fed into a fusion module. In fusion modules, both outputs are combined using Atrous Special Pyramid Pooling  \cite{Chen2016} module. The use of ASPP helps their model to preserve multi-scale contextual information. Finally, the Fusion module produced semantic region of objects with a refined boundary.  

The process of enlarging the receptive field using multi-resolution pyramid based representation helps the above model to incorporate scale information of objects to acquire fine-grained semantic segmentation. But capturing contextual information using receptive field enlargement may not be the only solution left for better semantic segmentation.

\subsection{Discussion}
From the year 2012, different CNN based semantic segmentation models have emerged in successive years to date. In subsection \ref{sem_seg}, we have described major up-gradation in the networks of various state-of-the-art models for better semantic segmentation. Among different models, Fully Convolutional Network (FCN) has set a path for semantic segmentation. Various models have used FCN as their base model. DeepLab and its versions have used atrous algorithm in different ways. SegNet, DeconvNet, U-Net have a similar architecture where the second part of those architectures is hierarchically opposite of the first half. ParseNet, PSPNet, and GCN have addressed semantic segmentation with respect to global contextual information. FC-DenseNet used top-down/bottom-up approach to incorporate low-level features with high-level features. So far, we have seen that the performance of a semantic segmentation model depends on the internal architecture of a network. In following subsections, we will see that it also depends on some other aspects such as the size of the data set, number of semantically annotated data,  different training hyperparameters, optimization algorithm, loss function, etc. We have shown those different comparative aspects of each model in tabular form. 
 
\subsubsection{Optimization details of different State-of-the-art Semantic Segmentation Models:} 

Table \ref{TD_SS} shows different optimization details of different models where we can see that the success of a model not only depends on the architecture. Comparison of different models with respect to optimization or training details shows that most of the researcher used stochastic gradient descent (SGD) as optimization algorithm but with different mini batch size of images. The choice of mini-batch size depends also on the number of GPU used to train a particular model.It is also shown here that most of the researcher have used momentum approximately same as 0.9. The main important feature in training a model is choosing the learning rate so that the model can converge in a optimized way. Regularization term is also an important factor in model training to combat overfitting. So, as network design, choosing appropriate hyperparameters is also a crucial thing in training to reach to a desirable accuracy. 
\begin{table}[htb!]
\caption{ Optimization details of different state-of-the-art semantic segmentation models}
\label{TD_SS}
\scriptsize
\begin{center}
\begin{tabularx}{\textwidth}{|X|X|X|p{3cm}|c|X|}
\hline
\textbf{Name of \newline the model }& \textbf{Optimization Algorithm} &\textbf{Mini \newline Batch Size}&\textbf{Learning Rate}&\textbf{Momentum}&\textbf{Weight \newline Decay}  \\ 
\hline 
FCN-VGG16 \cite{Long2017}& SGD \cite{Bottou2010}& 20 images& 0.0001 &0.9 &0.0016 or 0.0625 \\ \hline

DeepLab \cite{chen2014}&SGD &20 images & initially 0.001 (0.01 for final classification layer), increasing it by 0.1 at every 2000 iteration.&0.9&0.0005  \\ \hline 

Deconvnet \cite{Noh2015} &SGD&- & 0.01   &0.9  &0.0005\\ \hline

U-Net \cite{Ronneberger2015}&SGD & Single image &   &0.99  &$ $  \\ \hline 
DialatedNet \cite{YuK15}&SGD&14 images&0.001&0.9& -\\ \hline

ParseNet \cite{Liu2015} &SGD&     & $1e-9$  &0.9  &$ $ \\ \hline     
SegNet \cite{Badrinarayanan2015}&SGD&  12 images   & 0.1 & 0.9  &$ $  \\ \hline                 
GCN \cite{PengC2017}& SGD& Single image &  &0.99 &0.0005 \\ \hline    
PSPNet \cite{Zhao2016} &SGD&16 images     & `poly' learning rate with base learning rate of 0.01 and power to 0.9 &0.9  &0.0001  \\ \hline                       
FC-DenseNet103 \cite{JegouDVRB16}&SGD& &initially $1e-3$ with an exponential decay of 0.995& &$1e-4$\\\hline

EncNet \cite{Zhang2018}&SGD&  16 images& 0.001 with the power of 0.9& 0.9 &0.0001\\\hline

Gated-SCNN \cite{Towaki2019}&SGD&16 images&$1e-2$ with polynomial decay policy& 0&\\\hline                         
\end{tabularx}  
\end{center}
\end{table}

\begin{table}[htb!]
	\caption{Base Model, data preprocessing technique and loss functions of different state-of-the-art semantic segmentation models.}
	\label{TD_SS2}
	\tiny
	\begin{center}
		\begin{tabularx}{\linewidth}{|p{1.2cm}|X|X|X|}\hline
			
			\textbf{Name of \newline the model }  & \textbf{Base \newline Network} &\textbf{Data pre-processing}    &\textbf{Loss \newline Funtion}\\ \hline 
			
			FCN-VGG16 \cite{Long2017}&AlexNet\cite{Krizhevsky2012}, VGGnet\cite{ simonyan14}, GoogLeNet\cite{szegedy15} (All pre-trained on ILSVRC dataset \cite{ILSVRC15})& &Per-pixel multinomial logistic loss \\ \hline
			
			DeepLab \cite{chen2014}&VGG16 \cite{ simonyan14} pre-trained on ILSVRC dataset &Data augmentation using extra annotated data of \cite{Hariharan2011}&Sum of cross-entropy loss \\ \hline 
			
			Deconvnet  \cite{Noh2015} &VGG16 pre-trained on ILSVRC dataset&Data augmentation using extra annotated data of \cite{Hariharan2011} & \\ \hline
			
			U-Net \cite{Ronneberger2015}&FCN \cite{Long2017}&Data augmentation by applying random elastic deformation to the available training images&  Cross entropy loss   \\ \hline       
			
			DialateNet \cite{YuK15} & VGG16 \cite{ simonyan14}&Data augmentation using extra annotated data of \cite{Hariharan2011}&	\\ \hline	
			
			ParseNet  \cite{Liu2015}&FCN \cite{Long2017}&      &    \\ \hline    
			
			SegNet \cite{Badrinarayanan2015}  &VGG16  \cite{ simonyan14}&Local contrast normalization to RGB data       &Cross \newline entropy loss    \\ \hline                 
			
			GCN \cite{PengC2017}&ResNet152 \cite{he16} as feature network and FCN-4 \cite{Long2017} as segmentation network    &Semantic Boundaries Dataset \cite{Hariharan2011} is used as auxiliary dataset    &  \\ \hline    
			
			PSPNet \cite{Zhao2016} &Pretrained ResNet \cite{he16}    & Data augmentation: random mirror and random resize between 0.5 and 2, random rotation between -10 and 10 degrees, random Gaussian blur  & Four losses:\newline \textbullet{ Additional loss for initial result generation} \newline \textbullet{ Final loss for learning the residue later}\newline \textbullet{ Auxiliary loss for shallow layers}\newline \textbullet{ Master branch loss for final prediction}\\ \hline        
			
			FC-DenseNet \cite{JegouDVRB16} &DensNet  \cite{he16}  &Data augmentation using random cropping and vertical flipping &\\\hline
			
			EncNet \cite{Zhang2018} &ResNet & Data augmentation using random flipping, scaling, rotation and finally cropping & Semantic Encoding loss\\\hline
			
			Gated-SCNN \cite{Towaki2019} &ResNet101\cite{he16} and WideResNet\cite{ZagoruykoK16}&&\textbullet{ Segmentation loss for regular stream }\newline \textbullet{ Dual task loss for shape stream}\newline \textbullet\textbullet{ Standard binary cross entropy loss for boundary refinement} \newline \textbullet\textbullet{ Standard cross entropy for semantic segmentation} \\\hline  
		\end{tabularx}  
	\end{center}
\end{table}

\begin{table}[htb!]
	\caption{ Some important features of different state-of-the-art semantic segmentation models }
	\label{TD_SS3} 
	\centering
	\scriptsize
	\begin{tabularx}{\textwidth}{|p{2cm}|X| }\hline
		
		\textbf{Model} & \textbf{Important Features}  \\ \hline	
		
		FCN-VGG16 &\textbullet{ Dropout is used to reduce overfitting} \newline \textbullet{ End to end trainable}      \\ \hline
		
		DeepLab   &\textbullet{ End to end trainable}\newline  \textbullet{ Piecewise training for DCNN and CRF} \newline \textbullet{ Inference time during testing is 8frame per second}\newline \textbullet{ Used Atrous Special Pyramid Pooling module for aggregating multi-scale features}  \\ \hline
		
		Deconvnet  & \textbullet{ Used edge-box to generate region proposal}\newline \textbullet{ Used Batch Normalization to reduce internal covariate shift and removed dropout}\newline \textbullet{ Two-stage training for easy examples and for more challenging examples}\newline \textbullet { End to end trainable}\newline \textbullet { Drop-out layer is used at the end of the contracting path}  \\\hline    
		
		U-Net&\textbullet{ End to end trainable} \newline \textbullet{ Inference time for testing was less than 1 sec per image} \\ \hline              
		
		DialatedNet &     Two stage training: \newline \textbullet { Front end module with only dilated convolution} \newline \textbullet{ Dilated convolution with multi-scal context module } \\ \hline
		
		ParseNet &   \textbullet{ End to end trainable}\newline \textbullet{ Batch Normalization is used}\newline \textbullet{ Drop-out of 0.5 is used in deeper layers}   \\ \hline
		
		SegNet& \textbullet{ Different Ablation study} \\ \hline
		
		GCN & \textbullet{ Large Kernel Size} \newline \textbullet{ Included Global Contextual information}\\ \hline		
		
		PSPNet  &    \textbullet{ End to end training} \newline \textbullet{ Contains dialated convolution} \newline \textbullet{ Batch normalization} \newline \textbullet{ Used pyramid pooling module for aggregating multi-scale features} \\ \hline
		
		FC-DensNet& \textbullet{ Initialized the model with HeUniform\cite{HeZR015} and trained it with RMSprop dataset\cite{DauphinVCB15}} \newline \textbullet{ Used dropout of 0.2}\newline \textbullet{ Used the model parameters efficiently}\\\hline
		
		EncNet& \textbullet{ Used Context Encoding Module }\newline \textbullet{ Applied SE-loss to incorporate global semantic context }\\\hline
		
		Gated -SCNN& \textbullet{ End to end trainable} \newline \textbullet{ Applied ablation study }\\\hline
		\end{tabularx}
\end{table}

Table \ref{TD_SS2} presents base network (pre-trained on ImageNet \cite{Russakovsky2015} dataset), data pre-processing technique (basically data augmentation) and different loss function used for different models. Choosing of base network of a semantic segmentation model changes overtime according to the evolution of classification model. Optimization of a model always starts with some kind of data pre-processing. Different researcher have used different technique to pre-process the data. Most commonly used data preprocessing technique is data augmentation. As a loss function, cross entropy loss is used in most cases. Also, according to the complexity of model design, researchers have used different loss function to get higher accuracy. So, the choice of base network, data pre-processing technique, loss function etc are also very important to design a successful model. 

\begin{table}[h]
\caption{Comparative accuracy of different semantic segmentation models in terms of mean average precision (mAP) as Intersection over Union (IoU)}
\label{table4} 
\centering
	\scriptsize
\begin{tabularx}{\textwidth}{|l|c|X|X| }\hline

\textbf{Model} & \textbf{Year} & \textbf{Used Dataset} & \textbf{mAP  as IoU} \\	\hline
FCN-VGG16 \cite{Long2017} & 2014& Pascal VOC 2012 \cite{Everingham2007} &  62.2\%        \\ \hline

DeepLab\cite{chen2014}    & 2014& Pascal VOC 2012  &   71.6\%     \\ \hline

Deconvnet\cite{Noh2015}  &2015&Pascal VOC 2012   &  72.5\%     \\\hline    

U-Net\cite{Ronneberger2015}& 2015&ISBI cell tracking challenge 2015  & 92\%  on PhC-U373 and 77.5\% on DIC-HeLa dataset         \\ \hline              

DialatedNet \cite{YuK15}&2016 &Pascal VOC 2012 &73.9\% \\ \hline

ParseNet \cite{Liu2015}&2016 & \textbullet { ShiftFlow \cite{Liu2011}} \newline \textbullet { PASCAL- Context \cite{mottaghi2014}}\newline \textbullet { Pascal VOC 2012}   & 40.4\% \newline 36.64\% \newline  69.8\%            \\ \hline

SegNet \cite{Badrinarayanan2015}& 2016&\textbullet{ CamVid road scene segmentation \cite{BrostowFC09}} \newline \textbullet{ SUN RGB-D indoor scene segmentation\cite{SongLX15}}&60.10\% \newline \newline 31.84\%\\ \hline

GCN\cite{PengC2017} &2017 &\textbullet{ PASCAL VOC 2012} \newline \textbullet{Cityscapes \cite{Cordts2016}}      &  82.2\% \newline 76.9\%            \\ \hline		

PSPNet \cite{Zhao2016}  &2017 & \textbullet{ PASCAL VOC 2012} \newline \textbullet{ Cityscapes}  & 85.4\% \newline 80.2\%                \\ \hline

FC-DenseNet103 \cite{JegouDVRB16}&2017&\textbullet{ CamVid road scene segmentation } \newline \textbullet{ Gatech\cite{RazaGE15}}& 66.9\% \newline 79.4\% \\\hline

EncNet \cite{Zhang2018}&2018& \textbullet{ Pascal VOC 2012 } \newline \textbullet{ Pascal Context}& 85.9\% \newline 51.7\% \\\hline

Gated-SCNN \cite{Towaki2019}&2019&\textbullet{ Cityscapes}& 82.8\%\\\hline
\end{tabularx}
\end{table}
To give a clear view on the success of each model, we have listed some important features of each state-of-the-art model in table \ref{TD_SS3}.
	
	\subsubsection{Comparative Performance of State-of-the-art Semantic Segmentation Models:}
	In this section, we are going to show the comparative result of different state-of-the-art semantic segmentation models on various datasets in table \ref{table4}. The performance metric used here is mean average precision (mAP) as Intersection over Union (IoU) threshold. To have a better understanding, we have listed them in chronological order.
	
\section{Instance Segmentation}\label{is}
Like semantic segmentation, the applicability of CNN has been spread over instance segmentation too. Unlike semantic segmentation, instance segmentation masks each instance of an object contained in an image independently \cite{wu2016bridging, HariharanAGM14}. The task of object detection and instance segmentation are quite correlated. In object detection, researchers use the bounding box to detect each object instance of an image with a label for classification. Instance segmentation put this task one step forward and put a segmentation mask for each instance. 

Concurrent to semantic segmentation research, instance segmentation research has also started to use the convolutional neural network(CNN) for better segmentation accuracy. Herein, we are going to survey the evolution of CNN based instance segmentation models. In addition, we are going to bring up here an elaborate exploration of some state-of-the-art models for instance segmentation task. 
\subsection{Evolution of CNN based Instance Segmentation Models:}
 CNN based instance segmentation has also started its journey along with semantic segmentation.  As we have mentioned in section \ref{is} that instance segmentation task only adds a segmentation mask to the output of object detection task. That is why most of the CNN based instance segmentation models have used different CNN based object detection models to produce better segmentation accuracy and to reduce test time.

 Hariharan et al. have followed the architecture of R-CNN \cite{Girshick2014} object detector and proposed a novel architecture for instance segmentation called Simultaneous Detection and Segmentation(SDS) \cite{HariharanAGM14} which is a 4 step instance segmentation model as described in section \ref{is_bb}. 

 Till this time CNN based models have only used the last layer feature map for classification, detection and even for segmentation. In 2014, Hariharan et al. have again proposed a concept called Hyper-column \cite{Hariharan14a} which has used the information of some or all intermediate feature maps of a network for better instance segmentation. 
 The authors added the concept of Hyper-column to SDS and their modified network achieved better segmentation accuracy.

 Different object detector algorithms such as R-CNN, SPPnet \cite{He2014}, Fast R-CNN \cite{Girshick2015} have used two stages network for object detection. The first stage detects object proposals using Selective Search \cite{Uijlings2013} algorithm and second stage classify those proposals using different CNN based classifier.  Multibox   \cite{ErhanSTA13, SzegedyREA14}, Deepbox \cite{KuoHM15}, Edgebox \cite{Zitnick2014} have used CNN based proposal generation method for object detection. Faster R-CNN \cite{Ren2015} have used CNN based `region proposal network (RPN)' for generating box proposal.  However, the mode of all these proposal generations is using a bounding box and so the instance segmentation models.  In parallel to this, instance segmentation algorithms such as  SDS and Hyper-column have used Multi-scale Combinatorial Grouping (MCG)\cite{Arbelaez2014} for region proposal generation. DeepMask \cite{Pinheiro2015}, as discussed in section \ref{is_mask}, has also used CNN based RPN as Faster R-CNN to generate region proposals so that the model can be trained end to end. 

 Previous object detection and instance segmentation modules such as  \cite{Girshick2014},  \cite{He2014}, \cite{Girshick2015}, \cite{Ren2015} \cite{HariharanAGM14}, \cite{Hariharan14a}, \cite{Pinheiro2015} etc. have used computationally expensive external methods for generating object level or mask level proposals like Selective Search, MCG, CPMC \cite{Carreira2012}, RPN etc. Dai et al.  \cite{Dai2015} break the tradition of using a pipeline network and did not use any external mask proposal method. The authors have used a cascaded network for incorporating features from different CNN layers for instance segmentation. Also, the sharing of convolution features leads to faster segmentation models. Detail of the network is discussed in section \ref{is_bb}.  

 SDS, DeepMask, Hyper-columns have used feature maps from top layers of the network for object instance detection which leads to coarse object mask generation. Introduction of skip connection in \cite{Long2014, XieT15, Pierre2012, Zagoruyko16} reduces the coarseness of masks which is more helpful for semantic segmentation rather instance segmentation. Pinheiro et. al.\cite{Pinheiro2016} have used their model to generate a coarse feature map using CNN and then refined those models to get pixel-accurate instance segmentation masks using a refinement model as described in section \ref{is_mask}.

 In papers \cite{Torralba2003}, \cite{Semanet2012}, \cite{Szegedy2014}, \cite{He2014}, \cite{BellZBG15}, \cite{Hariharan14a}, \cite{Long2017}, \cite{Derek2012}, researchers used contextual information and low level features into CNN in various ways for better segmentation. Zagoruko et al.  \cite{Zagoruyko16} has also used those ideas by integrating skip connection, foveal structure and integral loss in Fast R-CNN \cite{Girshick2015} for better segmentation. Further description is given in section \ref{is_mpn}.
   
 Traditional CNNs are translation invariant i.e images with the same properties but with different contextual information will score the same classification score. Previous models, specially FCN, used a single score map for semantic segmentation. But for instance segmentation, a model must be translation variant so that the same image pixel of different instances having different contextual information can be segmented separately. Dai et al \cite{Dai2016} integrated the concept of relative position into FCN to distinguish multiple instances of an object by assembling a small set of score maps computed from the different relative positions of an object. Li et al \cite{LiQDJW16} extended the concept of  \cite{Dai2016} and introduced two different position-sensitive score maps as described in section \ref{is_rp}. 
   
 SDS, Hypercolumn, CFM \cite{DaiH014}, MNC \cite{Dai2015}, MultiPathNet\cite{Zagoruyko16} used two different subnetworks for object detection and segmentation which prevent the models to become an end to end trainable. On the other hand \cite{Liang2015},\cite{LiuS2016} extends instance segmentation by grouping or clustering FCNs score map which involves a large amount of post-processing.  \cite{LiQDJW16} introduced a joint formulation of classification and segmentation masking sub-networks in an efficient way.
   
 While \cite{ArnabT17, BaiU16, KirillovLASR16, LiuJSR2017} have used semantic segmentation models, Mask R-CNN \cite{He2017} extends the object detection model Faster R-CNN by adding a binary mask prediction branch for instance segmentation. In \cite{HuangZ2019}, Huang et al infused a network block in Mask R-CNN to learn the predicted mask in a qualitative way and proposed Mask Scoring R-CNN. Recently, Kirillov et al.\cite{Alex2019} used point-based rendering in Mask R-CNN and produce state-of-the-art instance segmentation model. 
   
 The authors of \cite{Uhrig2016}, \cite{Chen2017MaskLab} has introduced direction features to predict different instances of a particular object. \cite{Uhrig2016} has used template matching technique with direction feature to extract the center of an instance whereas \cite{Chen2017MaskLab} followed the assembling process of \cite{LiQDJW16, Dai2016} to get instances.

 The papers \cite{KongYCS16, ZagoruykoLLPGCD16, BaiU16, Hariharan14a} have used features form intermediate layers for better performance. Liu et al.\cite{Liu2018} have also used the concept of feature propagation from a lower level to top-level and built a state-of-the-art model based on Mask R-CNN as discussed in section \ref{is_fp}. In \cite{Newell2017}, Newell et al used a novel idea to use CNN with associative embedding for joint detection and grouping to handle instance segmentation.

 Object detection using the sliding window approach gave us quite successful work such as Faster R-CNN, Mask R-CNN, etc. with refinement step and SSD\cite{Liu2016}, RetinaNet\cite{Lin2017} without using refinement stage. Though sliding window approach is popular in object detection but it was missing in case of instance segmentation task. Chen et al. \cite{Chen2019} have introduced dense instance segmentation to fill this gap and introduced TensorMask.
 
\subsection{Some State-of-the-art Instance Segmentation Models:}\label{is_model} 
In this section, we are going to elaborately discuss architectural details of some state-of-the-art CNN based instance segmentation models. The models are categorized on the basis of the most impotant feature used. At the end of each categorical discussion, we have also briefly discussed the advantages and weaknesses of a particular model category in brief.
\subsubsection{Based on bounding box proposal generation:} \label{is_bb}

\textbf{SDS:} 
 Simultaneous Detection and Segmentation (SDS) \cite{HariharanAGM14} model consists of 4 steps for instance segmentation. The steps are proposal generation, feature extraction, region classification, and region refinement respectively.
\begin{figure}[htb]
	\centering
	\includegraphics[width=.95\linewidth]{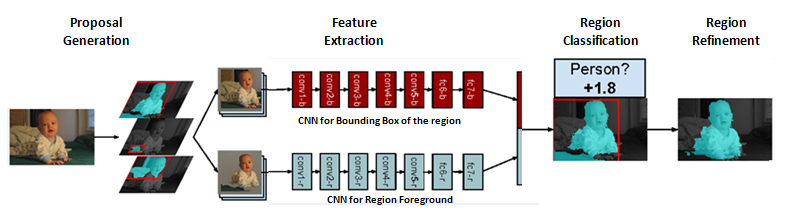}
	\caption{Architecture of SDS Network. From \cite{HariharanAGM14}}
	\label{sds}
\end{figure}
On input image, the authors have used Multi-scale Combinatorial Grouping(MCG) \cite{Arbelaez2014} algorithm for generating region proposals. Then each region proposals are fed into two CNN based sibling networks. As shown in figure \ref{sds}, the upper CNN generates a feature vector for bounding box of region proposals and the bottom CNN generates a feature vector for segmentation mask. Two feature vectors are then concatenated and class scores are predicted using SVM for each object candidate. Then non-maximum suppression is applied on the scored candidates to reduce the set of same category object candidates. Finally, to refine surviving candidates CNN feature maps are used for mask prediction.

\begin{figure}[htb]
	\centering
	\includegraphics[width=.94\linewidth]{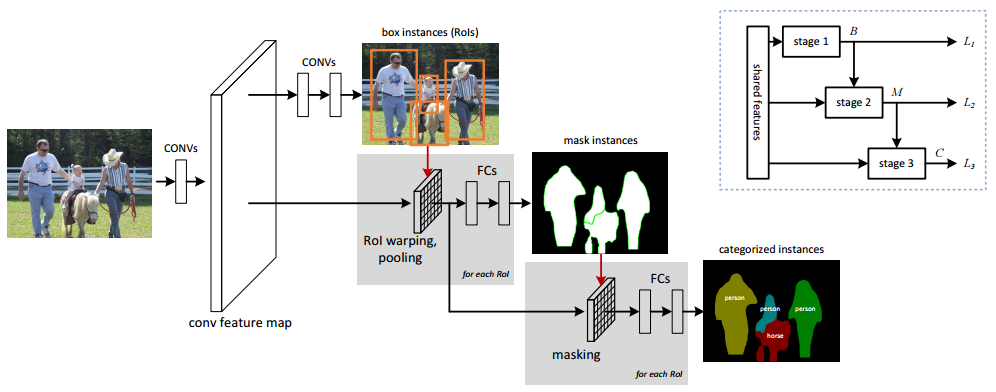}
	\caption{Three stage architecture of Multi-task Network Cascades. From \cite{Dai2015}.}
	\label{mnc}
\end{figure}

\textbf{Multi-task Network Cascades (MNC):}
Dai et al. \cite{Dai2015} used a network with the cascaded structure to share convolutional features and also used region proposal network (RPN) for better instance segmentation. The authors have decomposed the instance segmentation task into three sub tasks: instance differentiation (class agnostic bounding box generation for each instance), mask estimation (estimated a pixel-level mask/instance ) and object categorization (instances are labeled categorically). They proposed Multi-task Network Cascades (MNC)  to address these sub-tasks in three different cascaded stages to share convolutional features.  As shown in figure \ref{mnc}, MNC takes an arbitrary sized input which is a feature map extracted using VGG16 network. Then at the first stage, the network generates object instances from the output feature map as class agnostic bounding boxes with an objectness score using RPN. Shared convolutional features and output boxes of stage-1 then go to the second stage for regression of mask level class-agnostic instances. Again, shared convolutional features and output of the previous two stages are fed into the third stage for generating category score for each instance. 

\textbf{Mask R-CNN:}
Mask R-CNN\cite{He2017} contains three branches for predicting class, bounding-box and segmentation mask for instances within a region of interest (RoI). This model is the extension of Faster R-CNN. As Faster R-CNN, Mask R- CNN contains two stages. In the first stage, it uses RPN to generate RoIs. Then to preserve the spatial location, the authors have used RoIAlign instead of RoIPool as in Faster R-CNN. In the second stage, it simultaneously predicts a class label, a bounding box offset and a binary mask for each individual RoI. In Mask R-CNN, the prediction of binary mask for each class was independent and it was not a multi-class prediction.  
\begin{figure}[htb]
	\centering
	\includegraphics[width=.85\linewidth]{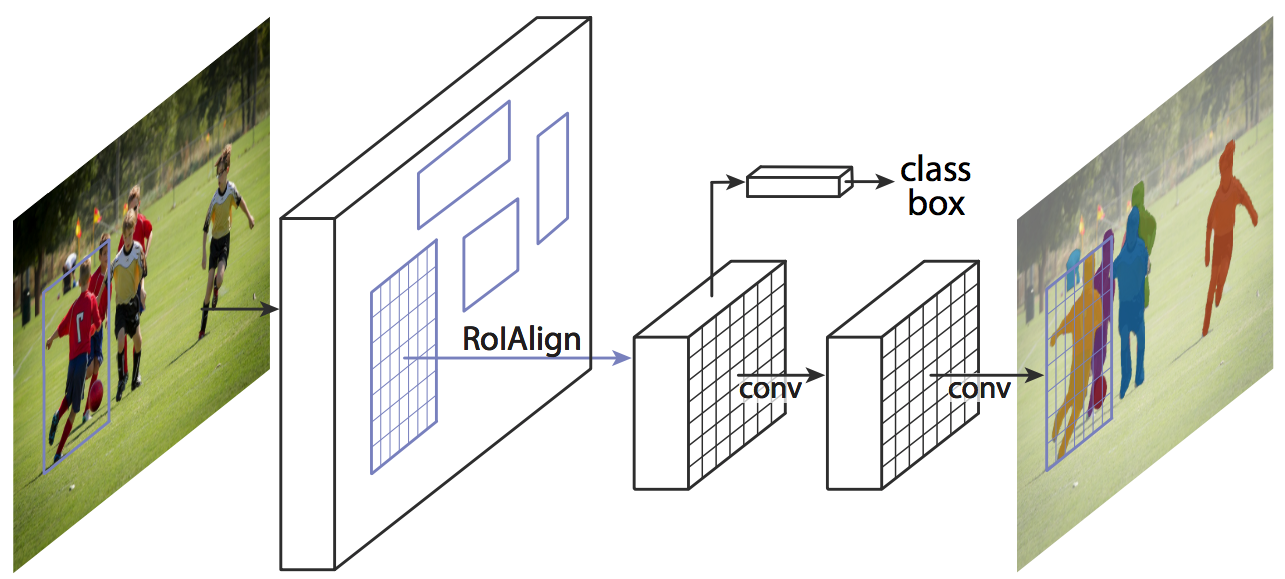}
	\caption{Architecture of  Mask R-CNN. From\cite{He2017}.}
	\label{mrcnn}
\end{figure}
 
 Generation of bounding box is computationally cost effective and its very helpful for detecting object. But its leads to computationally expensive alignment procedures. Also bounding box generation based models needs to generate masks for each instance separately. To overcome this problem researcher tried to generate segmentation mask proposal instead of bonding box proposal.

\subsubsection{Based on segmentation mask proposal generation:}\label{is_mask}
\textbf{DeepMask:}
 DeepMask \cite{Pinheiro2015} used CNN to generate segmentation proposals rather than less informative bounding box proposal algorithms such as Selective Search, MCG, etc. 
\begin{figure}[htb]
	\centering
	\includegraphics[width=.93\linewidth]{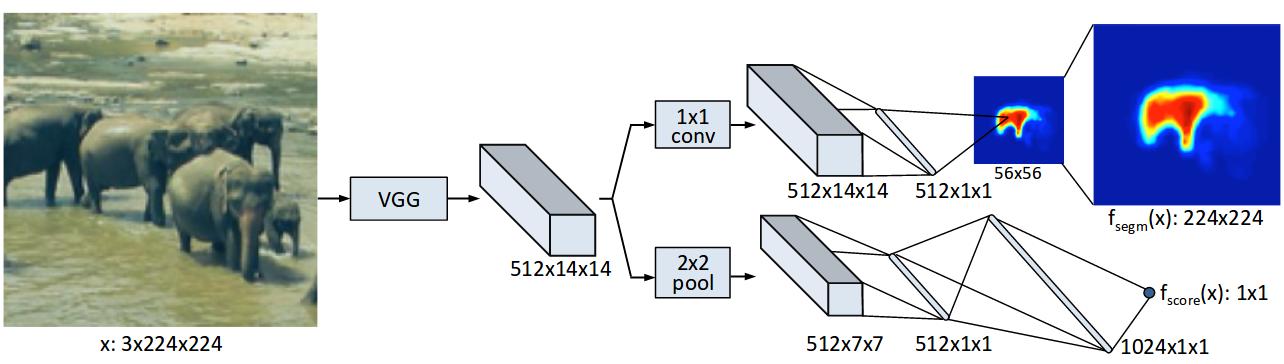} 
	\caption{Model illustration of DeepMask. From \cite{Pinheiro2015}.}
	\label{Deepmask}
\end{figure}  
DeepMask used VGG-A \cite{simonyan14} model (discarding last max-pooling layer and all fully connected layers) for feature extraction. As shown in figure \ref{Deepmask}, the feature maps are then fed into two sibling branches. The top branch which is the CNN based object proposal method of DeepMask predicts a class-agnostic segmentation mask and bottom branch assigns a score for estimating the likelihood of patch being centered on the full object. The parameters of the network are shared between the two branches.

\begin{figure}[htb]
	\centering
	\includegraphics[scale=0.6]{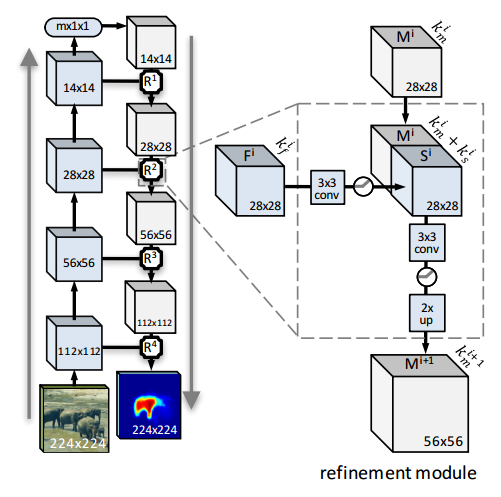}
	\caption{Bottom-up/top-down architecture of SharpMask. From \cite{Pinheiro2016}.}
	\label{sharpmask}
\end{figure}

\textbf{SharpMask:}
DeepMask generates accurate masks for object-level but the degree of alignment of the mask with the actual object boundary was not good.  SharpMask \cite{Pinheiro2016} contains a bottom-up feed-forward network for producing coarse semantic segmentation mask and a top-down network to refine those masks using a refinement module. The authors have used feed-forward DeepMask segmentation proposal network with their refinement module and named it as SharpMask. As shown in figure \ref{sharpmask}, the bottom-up CNN architecture produces coarse mask encoding. Then the output mask encoding is fed into a top-down architecture where a refinement module un-pool it using matching features from the bottom-up module. This process continues until the reconstruction of the full resolution image and the final object mask. 

Segmentation mask proposal generation using CNN helps the models to have better accuracy. But it does not have the power of capturing instances of object with different scales.

\subsubsection{Based on multi-scale feature incorporation:}\label{is_mpn}
\textbf{MultiPath Network:}
Zagoruko et al. integrate three modifications in the Fast R-CNN object detector and proposed Multipath Network \cite{Zagoruyko16} for both object detection and segmentation tasks. Three modifications are skip connections, foveal structure, and integral loss. Recognition of small objects without context is difficult. That is why, in \cite{Torralba2003}, \cite{Semanet2012}, \cite{Szegedy2014}, \cite{He2014}, \cite{GidarisK15}, the researcher used contextual information in various ways in CNN based model for better classification of objects. In Multipath Network, the authors have used four contextual regions called foveal regions. The view size of those regions are $1\times$, $1.5\times$,  $2\times$, $4\times$ of the original object proposal.
\begin{figure}[htb]
	\centering
	\includegraphics[width=.94\linewidth]{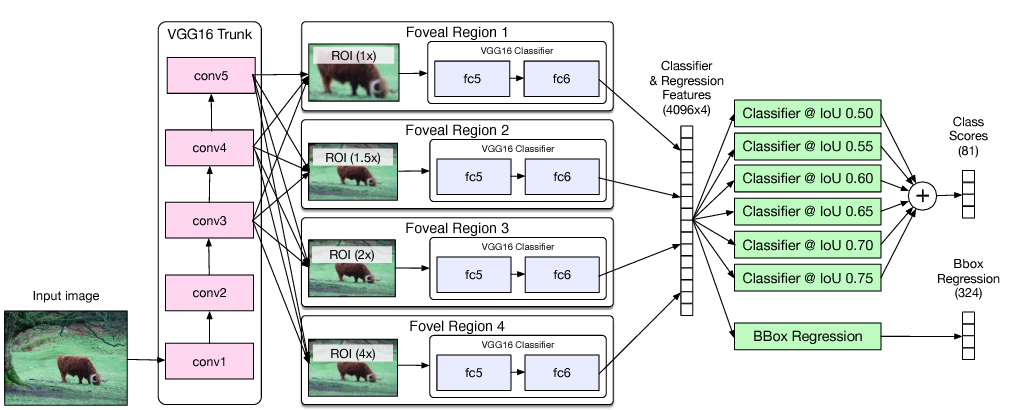}
	\caption{Architecture of  MultiPath Network. From \cite{Zagoruyko16}.}
	\label{mpn}
\end{figure}
 On the other hand, researchers of \cite{BellZBG15}, \cite{Hariharan14a}, \cite{Long2017}, \cite{Derek2012} has used feature from higher-resolution layers of CNN for effective localization of small objects. In Multipath Network, the authors have connected third, fourth and fifth convolutional layers of VGG16 to the four foveal regions to use multi-scale features for better object localization. Figure \ref{mpn} shows the architectural pipeline of MultiPath Network. Feature maps are extracted from an input image using the VGG16 network. Then using skip connection those feature maps go to four different Foveal Region. The output of those regions are concatenated for classification and bounding box regression. The use of the DeepMask segmentation proposal helped their model to be the 1st runner-up in MS COCO 2015 \cite{Lin2014} detection and segmentation challenges.

 This model tried to incorporate multi-scale feature maps to become scale invariant and also used skip connection to incorporate contextual information for better segmentation. But it lacks knowledge about relative position of an object instances.

\subsubsection{Based on capturing relative position of object instances:}\label{is_rp}
\begin{figure}[htb]
	\centering
	\includegraphics[scale=0.32]{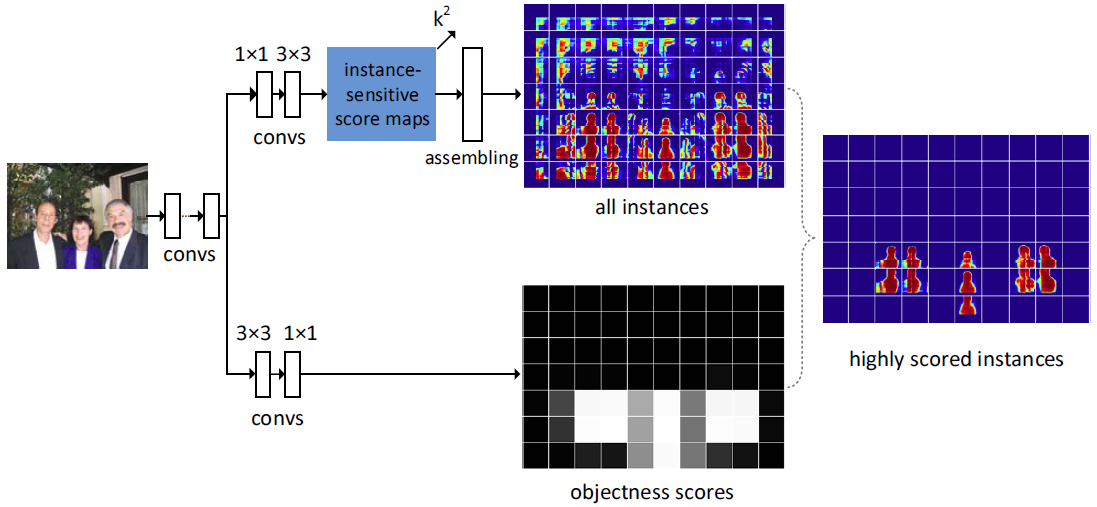}
	\caption{Architecture of Instance-sensitive fully convolutional network. From \cite{Dai2016}.}
	\label{ifcn}
\end{figure}

\textbf{InstanceFCN:}
The fully convolutional network (FCN) is good for single instance segmentation of an object category. But it can not distinguish multiple instances of an object. Dai et al have used the concept of relative position in FCN and proposed instance sensitive fully convolutional network (InstanceFCN) \cite{Dai2016} for instance segmentation. The relative position of an image is defined by a $k\times k$ grid on a square sliding window. This produces a set of $k^2$ instance sensitive score maps rather than one single score map as FCN. Then the instance sensitive score maps are assembled according to their relative position in a $m\times m$ sliding window to produce object instances. In DeepMask\cite{Pinheiro2015}, shifting sliding window for one stride leads to the generation of two different fully connected channels for the same pixel which is computationally exhaustive. In InstanceFCN, the authors have used the concept of local coherence \cite{Barnes2009} which means sliding a window does not require different computations for a single object. Figure \ref{ifcn} shows the architecture of InstanceFCN. 

\begin{figure}[htb]
	\centering
	\includegraphics[width=.93\linewidth]{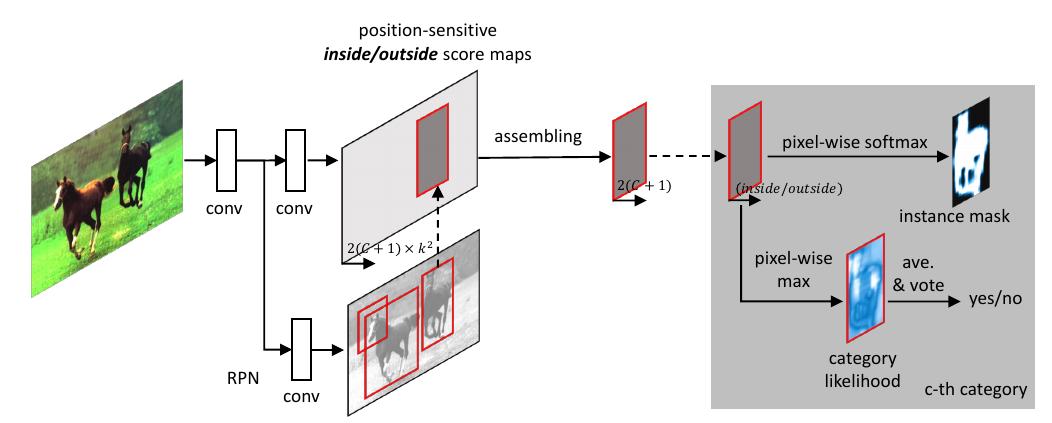}
	\caption{Architecture of  FCIs. From \cite{LiQDJW16}.}
	\label{fcis}
\end{figure}

\textbf{FCIs:} 
InstanceFCN introduced position-sensitive score mapping to signify the relative position of an object instance but the authors have used two different sub networks for object segmentation and detection. Because of two different networks, the solution was not end to end. Li et al. \cite{LiQDJW16} proposed the first end to end trainable fully convolutional network based model in which segmentation and detection are done jointly and concurrently in a single network by score map sharing as shown in figure \ref{fcis}. Also instead of the sliding window approach, the model used box proposals following \cite{Ren2015}. The authors have used two different position-sensitive score maps: position-sensitive inside score maps and position sensitive outside score maps. These two score maps depend on detection score and segmentation score of a pixel in a given region of interests (RoIs) with respect to different relative position. As shown in figure \ref{fcis} RPN is used to generate RoIs. Then RoIs are used on score maps to detect and segment object instances jointly.

\textbf{MaskLab:}
MaskLab \cite{Chen2017MaskLab} has utilized the merits of both semantic segmentation and object detection to handle instance segmentation. The authors have used Faster R-CNN\cite{Ren2015} (ResNet-101\cite{he16} based) for predicting bounding boxes for object instances. Then they have calculated semantic segmentation score maps for labeling each pixel semantically and direction score maps for predicting individual pixels direction towards the center of its corresponding instance. Those score maps are cropped and concatenated for predicting a coarse mask for target instance. The mask is then again concatenated with hyper-column features\cite{Hariharan14a} extracted from low layers of ResNet-101 and processed using a small CNN of three layers for further refinement. 
\begin{figure}[htb]
	\centering
	\includegraphics[width=.93\linewidth]{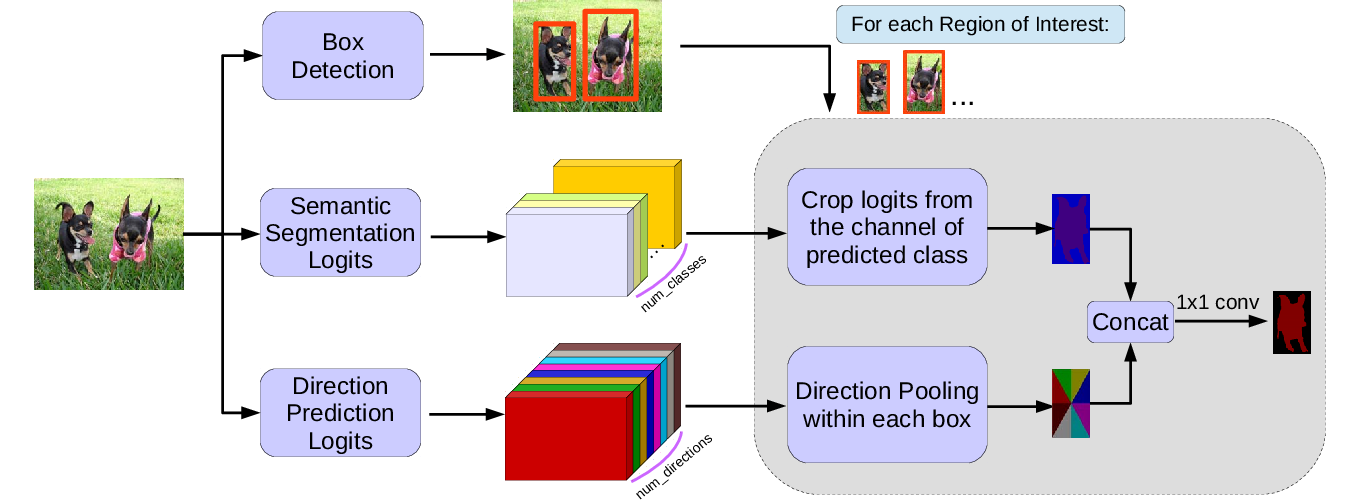}
	\caption{Architecture of  MaskLab. From \cite{Chen2017MaskLab}.}
	\label{mlab}
\end{figure}

Using position sensitive score maps the above models tried to capture the relative position of object instances.

\subsubsection{Based on feature propagation:}\label{is_fp}

\textbf{PANet:} 
The flow of information in the convolutional neural network is very important as the low-level feature maps are information-rich in terms of localization and the high-level feature maps are rich in semantic information. Liu et al. focused on this idea. Based on  Mask R-CNN and Feature Pyramid Network(FPN) \cite{Lin2016}, they have proposed a Path Aggregation Network (PANet) \cite{Liu2018} for instance segmentation. PANet used FPN as its base network to extract features from different layers. To propagate the low layer feature through the network, a bottom-up augmented path is used. Output of each layer is generated using previous layers high-resolution feature map and a coarse map from FPN using a lateral connection. Then an adaptive pooling layer is used to aggregate features from all levels. In this layer, a RoIPooling layer is used to pool features from each pyramid level and element wise max or sum operation is used to fuse the features. As Mask R-CNN, the output of the feature pooling layer goes to three branches for prediction of the bounding box, prediction of the object class and prediction of the binary pixel mask. 

Using feature propagation network and pooling pyramid, this model incorporates low to high level feature as well as multi-scale features which leads to information rich instance segmentation.
\begin{figure}[htb]
	\centering
	\includegraphics[scale=0.29]{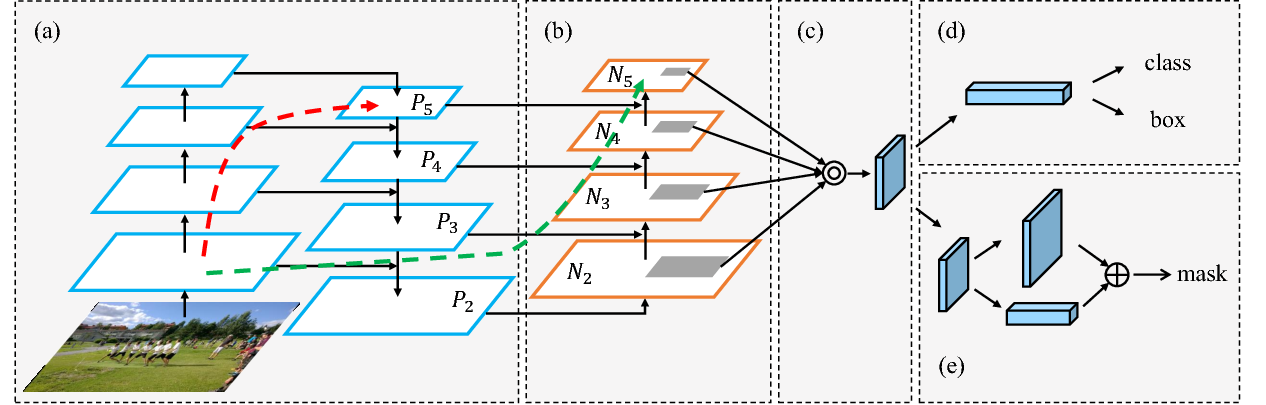}
	\caption{Architecture of PANet. From \cite{Liu2018}.}
	\label{panet}
\end{figure}

\subsubsection{Based on sliding window approach:}\label{is_sw}

\textbf{TensorMask:}
Previous instance segmentation models used methods in which the objects are detected using bounding box then segmentation is done. Chen et al. have used the dense sliding window approach instead of detecting the object in a bounding box named TensorMask \cite{Chen2019}. The main concept of this architecture is the use of structured high-dimensional (4D) tensors to present mask over an object region.  A 4D tensor is a quadruple of  (V, U, H, W). The geometric sub-tensor (H, W) represents object position and (V, U) represents the relative mask position of an object instance. Like feature pyramid network, TensorMask has also developed a pyramid structure, called $tensor bipyramid$ over a scale-indexed list of 4D tensors to acquire the benefits of multi-scale. 

\subsection{Discussion:}
In the previous subsection \ref{is_model}, we have presented important architectural details of different state-of-the-art models. Among them, some models are based on different object detection models such as R-CNN, Fast R-CNN, Faster R-CNN, etc. Some models are based on semantic segmentation models such as FCN, U-Net, etc. SDS, DeepMask, SharpMask are based on proposal generation. InstanceFCN, FCIs, MaskLab calculate position-sensitive score maps for instance segmentation. PANet emphasized on feature propagation across the network. TensorMask used the sliding window approach for dense instance segmentation. So, architectural differences help different models to achieve success in various instance segmentation dataset. On the other hand, fine-tuning of hyper-parameters, data pre-processing methods, choice of the loss function and optimization function, etc are also played an important role in the success of a model. In this subsections, we are going to present some of those important features in a comparative manner.  
\subsubsection{Optimization Details of State-of-the-art Instance  Segmentation Models:} The training and optimization process is very crucial for a model to become successful. Most of the state-of-the-art models used stochastic gradient descent(SGD) \cite{Goyal2017} as an optimization algorithm with different initialization of corresponding hyper parameters such as mini-batch size, learning rate, weight decay, momentum etc. Table \ref{TD_IS} shows those hyper-parameters in a comparative way. As semantic segmentation, most of the instance segmentation models  used momentum of 0.9 with different weight decay. Variation of choosing a learning rate is also not much.

\begin{table}[htb!]
	\caption{Optimization details of different state-of-the-art instance segmentation models}
	\label{TD_IS}
	\tiny
	\begin{center}
		\begin{tabularx}{\textwidth}{|X|X|X|p{3cm}|c|X|}\hline
			
			\textbf{Name of \newline the model }& \textbf{Optmization Algorithm} &\textbf{Mini \newline Batch Size}&\textbf{Learning Rate}&\textbf{Momentum}&\textbf{Weight \newline Decay}  \\ \hline	
			
			
			DeepMask \cite{Pinheiro2015}&SGD&32 images&0.001&0.9&0.00005\\ \hline
			
			MNC \cite{Dai2015}&SGD&1 images per GPU, total 8 GPUs are used&0.001 for 32k iteration, 0.0001 for next 8K iteration && \\\hline			
			MultPath Network \cite{Zagoruyko16}&SGD&4 images,1 image per GPU, each with 64 object proposals&initially 0.001, after 160k iterations, it was reduced to 0.0001&-&-\\\hline			
			SharpMask  \cite{Pinheiro2016}&SGD&&$1e^{-3}$&&\\\hline			
			InstanceFCN \cite{Dai2016}&SGD&8 images each with 256 sampled windows, 1 image/GPU&0.001 for initial 32k iterations and 0.0001 for the next 8k.&0.9&0.0005 \\\hline			
			FCIs \cite{LiQDJW16}&SGD&8 images/batch, 1 image per GPU&0.001 for the first 20k and 0.0001 for the next 10k iterations&& \\\hline			
			Mask R-CNN \cite{He2017}&SGD& 16 images/batch, 2 images per GPU &0.02 for first 160k iteration and 0.002 for next 120k iterations&0.0001& 0.9\\\hline			
			PANet \cite{Liu2018}&SGD&16 images&0.02 for 120k iterations and 0.002 for 40k iterations& 0.0001&0.9\\\hline			
			TensorMask \cite{Chen2019}&SGD&16 images, 2 images per GPU &0.02 with linear warm-up\cite{Goyal2017} of 1k iteration &0.9&0.0005 \\\hline			
		\end{tabularx}
	\end{center}
\end{table}

\begin{table}[htb!]
	\caption{Base Model, data preprocessing technique and loss functions of different Stat-of-the-art models.}
	\label{TD_IS2}
	\tiny
	\begin{center}
		\begin{tabularx}{\linewidth}{|p{1.2cm}|X|X|X|}\hline
			
			\textbf{Name of \newline the model }  & \textbf{Base \newline Network} &\textbf{Data pre-processing}    &\textbf{Loss \newline Funtion}\\ \hline
			
			SDS &&& \\\hline
			
			DeepMask&VGG-A pretrained on ImageNet dataset&\textbullet{ Randomly jitter `canonical' positive examples for increasing the model's robustness} \newline\textbullet{ Applied translation shift, scale deformation and horizontal flip for data augmentation}&Sum of binary logistic regression losses\newline \textbullet{ One for each location of the segmentation network}\newline\textbullet{ Other for the objectness score}\\\hline
			
			MNC &\textbullet{ VGG-16}\newline \textbullet{ ResNet-101}&&Unified loss function\newline\textbullet{ RPN loss for box regression/instance}\newline \textbullet{ Mask regression loss/instance}\newline\textbullet{ loss function for categorizing instances}\newline\textbullet{ Inference time per image is 1.4sec}\\\hline
			
			MultPath Network & Fast R-CNN&Horizontal flip as data augmentation&Integral loss function: \newline\textbullet{ Integral log loss function for classification }\newline\textbullet{ Bounding box function}\\\hline
			
			SharpMask &DeepMask&&Same as used in DeepMask \\\hline
			
			InstanceFCN &VGG-16, pretrained on ImageNet&Arbitray sized images are used for training with scale jittering following \cite{He2014}&Logistic regression loss for predicting abjectness score and segment instances \\\hline
			
			FCIs &ResNet-101&& \\\hline
			
			Mask R-CNN &Faster R-CNN based on ResNet &&Multi-task loss:
			\newline\textbullet{ log loss function for classification }\newline\textbullet{ L1 loss function for bounding box regression }\newline\textbullet{ Average binary cross entropy loss for mask prediction} \\\hline
			
			MaskLab \cite{Chen2017MaskLab}&ResNet-101 based Faster R-CNN pre-trained on ImageNet&& \\\hline
			
			PANet &ResNet-50, ResNeXt-101\cite{XieGDTH16} based Mask R-CNN and  FPN && \\\hline
			
			TensorMask &ResNet-50, FPN&Scale jittering is used& Weighted sum of all task loss specially for mask, per-pixel binary classification loss is used. Focal loss is used to handle foreground background class imbalance.  \\\hline				
		\end{tabularx}
	\end{center}
\end{table}

\begin{table}[htb!]
	\caption{ Some important features of different state-of-the-art instance segmentation models }
	\label{TD_IS3} 
	\centering
	\scriptsize
	\begin{tabularx}{\textwidth}{|p{2cm}|X| }\hline
		
		\textbf{Model} & \textbf{Important Features}  \\ \hline	
		
		SDS &\textbullet{ Used MCG to generate region proposals}\newline \textbullet{ Used segmentation data from SBD\cite{Hariharan2011}} \\\hline
		
		DeepMask&\textbullet{ The inference time in MS COCO is 1.6s per image} \newline\textbullet{ The inference time in PASCAL VOC 2007 is 1.2s per image}\ \\\hline
		
		MNC&\textbullet{ End to End trainable}\newline\textbullet{ Convolutional feature sharing leads to reduction of test time of 360ms/image. }  \\\hline
		
		Multi-path Network &\textbullet{ Skip Connection for sharing feature among multiple levels}\newline\textbullet{ Foveal structure to capture multi-scale object}\newline\textbullet{ Integrated loss function for improving localization }\newline\textbullet{ DeepMask region proposal algorithm to generate region proposals}\newline \textbullet{ Training time 500ms/image}\\\hline
		
		SharpMask & \textbullet{ Bottom-up/top-down approach}\newline\textbullet{ DeepMask used in bottom-up network to generate object proposal} \newline\textbullet{ Top-down network is stack of refinement model which aggregate features from corresponding layers from bottom-up path } \newline\textbullet{ Two stage training: One for bottom-up and another for top-down network}\\\hline
		
		InstanceFCN &\textbullet{ A small set of score maps computed from different relative position of an image patch are assembled for predicting the segmentation mask}\newline\textbullet{ Applied `hole algorithm\cite{chen2014}' in last three layers of VGGNet } \\\hline
		
		FCIs &\textbullet{ End to end trainable FCN based model}\newline\textbullet{ Based on position sensitive inside and outside score map}\newline\textbullet{ Inference time 0.24 seconds/image}\newline\textbullet{ Six times faster than MNC} \\\hline
		
		Mask R-CNN &\textbullet{  RoIAlign layers are used instead of RoIPool layer to preserve special location}\newline\textbullet{ Inference time was 200 ms per frame } \\\hline
		
		MaskLab & \textbullet{ Used atrous convolution to extract denser feature map to control output resolution}\newline\textbullet{ End to end trainable model}\newline\textbullet{ To cover 360 degree of an instance 8 directions are used with 4 number of distance quantization bins for direction pooling} \\\hline
		
		PANet & \textbullet{ FPN is used as Backbone network}\newline\textbullet{ Adaptive feature pooling layer is introduced}\\\hline
		
		TensorMask&\textbullet{ Dense instance segmentation using sliding window approach}\newline\textbullet{ The model works on 4D tensor} \\\hline
		
	\end{tabularx}
\end{table}	

Different models have used different CNN based classification, Object detection and semantic segmentation model as their base network according to the availability. It is an open choice to the researchers to choose a base model (may be pre-trained on some dataset) according to their application domain. Most of the data preprocessing basically includes different data augmentation technique. Differences in loss function depend on the variation of  the model architecture as shown in table \ref{TD_IS2}.  Table \ref{TD_IS3} is showing some important features of different models.

\begin{table*}[htb!]
	\centering
	\caption{Comparison of different instance segmentation models as average precision according to IoU threshold  }
	\label{C_IS} 
	\tiny
	\begin{tabularx}{\textwidth}{|c|c|X|X| }
		\hline
		\textbf{Model} & \textbf{Year} & \textbf{Used Dataset} & \textbf{mAP as IoU} \\	
		\hline
		SDS &2014&\textbullet{ PASCAL VOC 2011} $\longrightarrow$\newline \textbullet{ PASCAL VOC 2012} $\longrightarrow$&51.6\% \newline 52.6\% \\ \hline
		
		DeepMask &2015&\textbullet{ PASCAL VOC}\newline\textbullet{ MS COCO}&Fast R-CNN using DeepMask outperforms original Fast R-CNN and achieved 66.9\% accuracy on PASCAL VOC 2007 test dataset\\\hline
		
		MNC&2016 &\textbullet{ PASCAL VOC 2012} $\longrightarrow$\newline \textbullet{ MS COCO 2015} $\longrightarrow$&63.5\% on validation set\newline 39.7\%  on $test-dev set$ \newline (both mAP@).5 IoU threshold) \\\hline
		
		MultPath Network &2015&MS COCO 2015 $\longrightarrow$&25.0\%(AP), 45.4\%($AP^{50}$) and 24.5\% ($AP^{75})$, all on test dataset.\newline Superscripts of AP denotes IoU threshold\\\hline
		
		SharpMask +MPN \cite{Zagoruyko16}&2016&MS COCO 2015 $\longrightarrow$&25.1\%(AP), 45.8\%($AP^{50}$) and 24.8\% ($AP^{75}$), all on test dataset. Superscripts of AP denotes IoU threshold\\\hline
		
		InstanceFCN+MNC &2016&PASCAL VOC 2012 validation dataset  & 61.5\%(mAP@0.5) and 43.0\%(mAP@0.7) \\\hline
		
		FCIs &2017&\textbullet{ Pascal voc 2012 $\longrightarrow$}\newline\textbullet{ MS COCO 2016$\longrightarrow$} & 65.7\%(mAP@0.5) and 52.1\%(mAP@0.7)  59.9\%(mAP@0.5)(ensemble)\\\hline
		
		Mask R-CNN &2017&MS COCO&60.0\%($AP^{50}$) and 39.4\% ($AP^{75}$) , all on test dataset.Superscripts of AP denotes IoU threshold \\\hline
		
		MaskLab&2018&MS COCO (test-dev) $\longrightarrow$& 61.1\%(mAP@0.5) and 40.4\%(mAP@0.7)   \\\hline
		
		PANet &2018  &\textbullet{ MS COCO 2016 $\longrightarrow$}\newline\textbullet{ MS COCO 2017$\longrightarrow$}&65.1\%($AP^{50}$) and 45.7\% ($AP^{75}$)\newline69.5\%($AP^{50}$) and 51.3\% ($AP^{75}$) , (Mask AP).Superscripts of AP denotes IoU threshold  \\\hline
		
		TensorMask &2019&MS COCO (test-dev) & 37.3\% (AP), 59.5\%($AP^{50}$) and 39.5\% ($AP^{75}$).Superscripts of AP denotes IoU threshold\\\hline
	\end{tabularx}
\end{table*}

\subsubsection{Comparative Performance of State-of-the-art Instance Segmentation Models:} Around 2014, concurrent with semantic segmentation task, CNN based instance segmentation models have also started gaining better accuracy in various data sets such as PASCAL VOC, MS COCO, etc. In table \ref{C_IS}, we have shown the comparative performance of various state-of-the-art instance segmentation models on those datasets in chronological order.

\section{Panoptic Segmentation} Panoptic segmentation (PS)  \cite{kirillov2019panoptic, Daan2019, Liu2019, AKirillov2019, Xiong2019, Petrovai2019} is the combination of semantic segmentation and instance segmentation. This is a new research area these days. In this task, we need to associate all the pixels in the image with a semantic label for classification and also identify the instances of a particular class. The output of a panoptic segmentation model will contain two channels: one for pixel's label (semantic segmentation) and another for predicting each pixel instance (instance segmentation). 
In the following paragraphs we have given some examples of recently designed panoptic segmentation models.
\subsection{Some popular state-of-the-art panoptic segmentation Models}

Krillov et al. first proposed panoptic segmentation \cite{kirillov2019panoptic} by unifying semantic segmentation and instance segmentation to encompass both stuff and thing classes. The have used novel panoptic quality (PQ) metric to measure the performance of the segmentation. Their model produces simple but general output. The authors have used their model on Cityscapes, ADE20K \cite{Zhou2017}and Mapillary Vista \cite{Neuhold2017} datasets and got better accuracy on segmentation.

\textbf{OANet: }Liu et al. proposed end-to-end Occlusion Aware Network(OANet) \cite{Liu2019} for panoptic segmentation. The authors have used Feature Pyramid Network to extract feature maps from the input image. Upon extracted feature they have applied two different branches: One for semantic segmentation and another for instance segmentation. Mask R-CNN is used for instance segmentation branch. Output of both branches are fed into novel Spatial Ranking Module for final output of panoptic segmentation. They have applied their model on COCO panoptic segmentation benchmark and got promising results.   
\begin{figure}[htbp]
	\centering
	\includegraphics[scale=0.50]{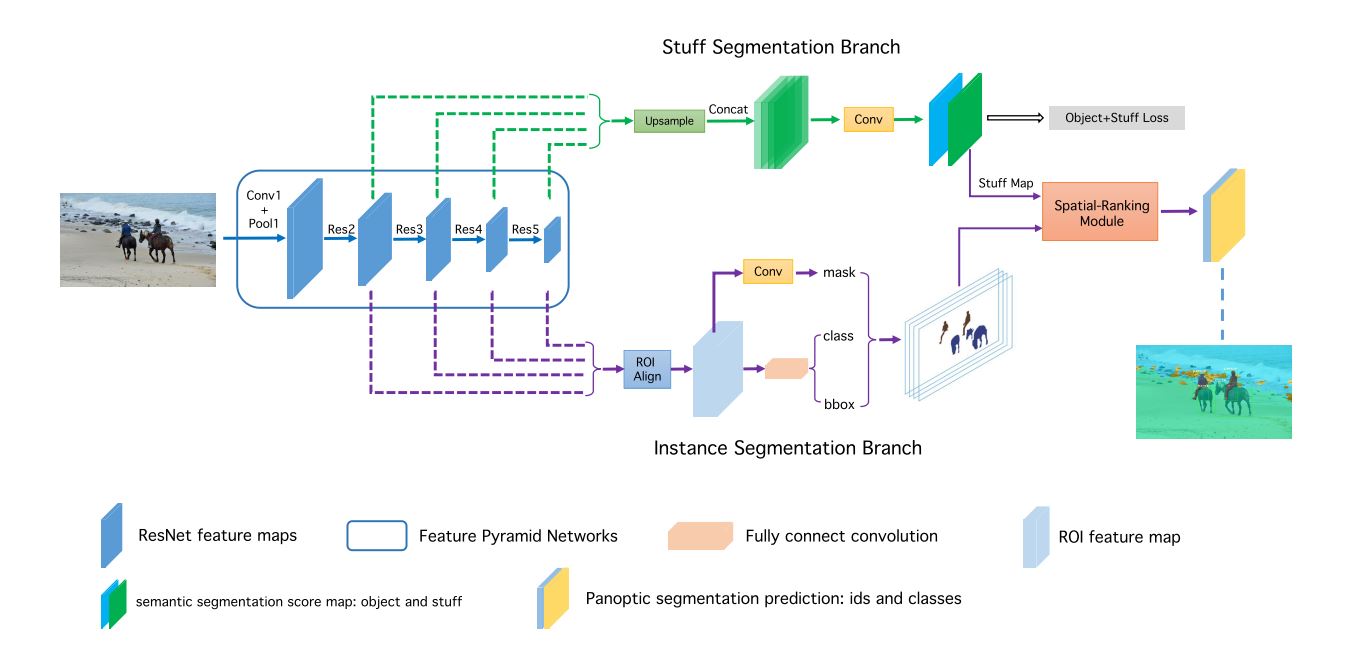}
	\caption{Architecture of Occlusion Aware Network(OANet). From \cite{Liu2019}}
	\label{OANet}
\end{figure}

\textbf{UPSNet:} Xiong et al. proposed a unified panoptic segmentation network (UPSNet) \cite{Xiong2019} to handle panoptic segmentation. The authors have used ResNet and FPN based Mask R-CNN as a backbone network to extract convolutional feature map. Those convolutional feature maps are fed into three sub-networks: for Semantic segmentation,for instance segmentation and for panoptic segmentation. Smantic segmentation sub-network consists of deformable convolutional network \cite{DaiJHYYGHY2017} for segmenting staff classes. Instance segmentation sub-network consists of three branch for bounding box regression, classification and sementation mask. All the outputs from these two subnetwork further goes to the panoptic segmentation sub-network for final panoptic segmentation. The authors have used teir model on Cityscapes and COCO datasets.
\begin{figure}[htbp]
	\centering
	\includegraphics[scale=0.60]{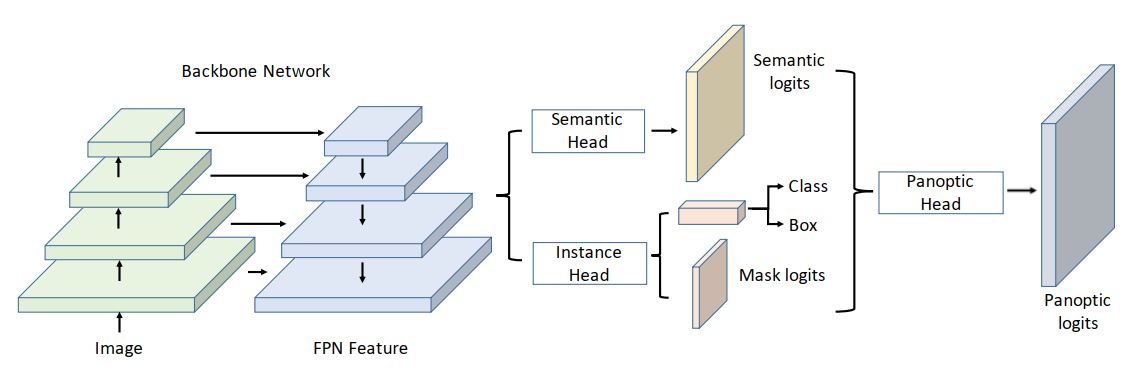}
	\caption{Architecture of unified panoptic segmentation network (UPSNet). From \cite{Xiong2019}}
	\label{UPSNet}
\end{figure}

\textbf{Multitask Network for Panoptic Segmentation:}Andra Petrovai and Sergiu Nedevschi have proposed an end to end trainable multi-task network \cite{Petrovai2019} for panoptic Segentation with the capability of object occlusion and scene depth ordering. As \cite{Xiong2019}, the authors have used ResNet and FPN based back bone network for multi-scale feature extraction. the output of backbone network fed into 4 individual sub-networks for four tasks. First sub-network is for object detection and classification using Faster R-CNN, Second one is for instance segmentation using Mask R-CNN, third one is for semantic segmentation using pyramid pooling module as used in PSPNet \cite{Zhao2016} and fourth one is for panoptic segmentation. The authors have used their model on Cityscapes dataset and got 75.4\% mIoU and 57.3\% PQ.

\begin{figure}[!htb]
	\centering
	\includegraphics[scale=0.40]{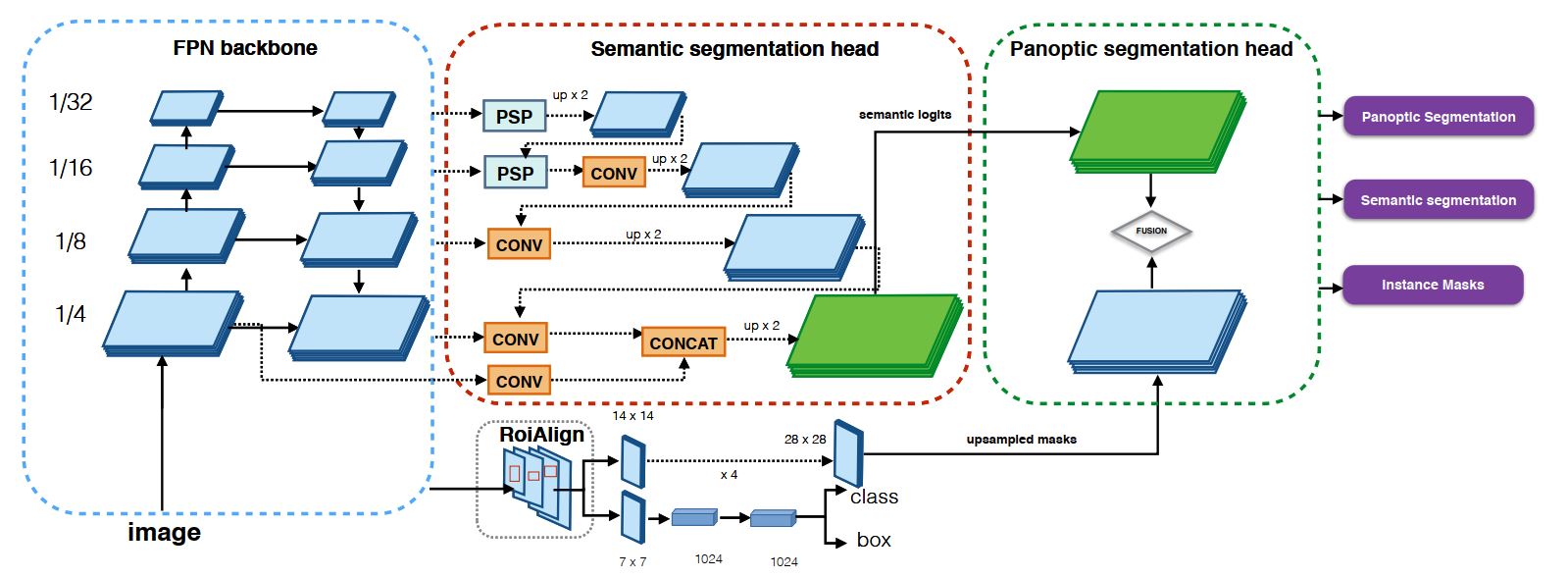}
	\caption{Architecture of Multitask Network for Panoptic Segmentation. From \cite{Petrovai2019}}
	\label{Mtnps}
\end{figure}
\section{Conclusion}
Image segmentation is a challenging work as it needs spatially variant features to preserve the context of a pixel for semantic labeling. Semantic segmentation categorizes each pixel with a semantic label whereas instance segmentation segments individual instances of objects contained in an image. The success of recent state-of-the-art models depends mostly on different network architecture. Except for that, various other aspects such as choice of the optimization algorithm, the value of hyper-parameters, data-preprocessing technique, choice of the loss function, etc are also responsible for becoming a successful model. In our article, we have presented the evolution of Convolutional Neural Networks based image segmentation models. We have categorically explored some state-of-the-art segmentation models along with their optimization details, and a comparison among the performance of those models on different datasets. Lastly, we have given a glimpse of recent state-of-the-art panoptic segmentation models. The application area of image segmentation is vast. According to the requirement of the application task, a suitable model can be applied using some domain-specific fine-tuning using dataset.  Overall this article gives systematic ideas about present state-of-the-art image segmentation that will help researchers of this area for further proceedings. 

\section*{Acknowledgment}
The authors of the article are thankful to the editor and anonymous reviewers for mentioning some comments and suggestions on initial submission. For such comments and suggestions, the article got this present shape. 
\bibliographystyle{elsarticle-num}
\footnotesize{\bibliography{objdbib}}

\end{document}